\documentclass[letterpaper, 10 pt, conference]{ieeeconf}  

\IEEEoverridecommandlockouts                              

\usepackage{amsmath} 
\usepackage{amssymb}  
\usepackage{graphicx} 
\usepackage{subcaption}
\usepackage{xurl}
\usepackage[textsize=tiny]{todonotes}
\usepackage{ifthen}
\usepackage[letterpaper, top=1in, bottom=0.75in, left=0.75in, right=0.75in]{geometry}
\newboolean{comments}
\setboolean{comments}{false} 

\ifthenelse{\boolean{comments}}{
  \setlength{\marginparwidth}{1.5cm}
  \newcommand{\zkn}[1]{\todo[color=orange!20, size=\tiny]{Zsolt: #1}}
  \newcommand{\zk}[1]{\textcolor{blue}{Zsolt: #1}}
  \newcommand{\gc}[1]{\todo[color=orange!20, size=\tiny]{GC: #1}}
}{
  \newcommand{\zk}[1]{}
  \newcommand{\zkn}[1]{}
  \newcommand{\gc}[1]{}
}

\usepackage{pifont}
\usepackage{xcolor}
\usepackage{array}

\newcommand{\cmark}{\textcolor{green!70!black}{\ding{51}}}
\newcommand{\xmark}{\textcolor{red}{\ding{55}}}

\title{\LARGE \bf
Splat2Act: From Captured Scenes to Personalized VLA Manipulation
}

\author{%
Xiaomeng Ye\textsuperscript{1}, 
Krishna Jaganathan\textsuperscript{1}, 
Gyubeum Lim\textsuperscript{1},\\
Gunjan Chhablani\textsuperscript{2}, 
Muhammad Zubair Irshad\textsuperscript{3}, 
Zsolt Kira\textsuperscript{1}%
\thanks{\textsuperscript{1}Institute for Robotics and Intelligent Machines, Georgia Institute of Technology, Atlanta, GA, USA. \textsuperscript{2}Waymo LLC, Mountain View, CA, USA. \textsuperscript{3}Toyota Research Institute, Cambridge, MA, USA. Emails: \{xye87, kjaganathan7, glim36, zkira\}@gatech.edu, \{chhablani.gunjan, muhammadzubairirshad\}@gmail.com.}
}

\newcommand{\shorthand}{Splat2Act}

\begin{document}
\bstctlcite{BSTcontrol}

\maketitle
\thispagestyle{empty}
\pagestyle{empty}

\begin{abstract}
A user who brings a generalist manipulation policy into their home or office wants it to pick up their objects from their shelves, counters, and tables. Yet even a strong generalist like $\pi_{0.5}$ achieves 0\% pickup success in our captured deployment scenes without adaptation, and collecting real demonstrations in every user's environment is prohibitively expensive. We present \shorthand{}, an automated real-to-sim-to-real pipeline that personalizes a pretrained vision-language-action (VLA) policy to a specific scene and object set using only a handheld phone scan and images of the user's objects. Unlike prior real-to-sim work that reconstructs individual tabletops, \shorthand{} operates on building-scale indoor environments: it builds a photorealistic digital twin with 3D Gaussian Splatting and collision geometry, automatically discovers tabletop and non-tabletop receptacles, constructs feasible pickup tasks, and generates successful demonstrations with a privileged simulation oracle, with only scene and object capture required from a human. Applied to 101 scanned campus environments with a median navigable area of 104.87 m$^2$ and 7 receptacles, the pipeline yields 852 receptacles and over 68,000 pickup demonstrations. We fine-tune $\pi_{0.5}$ in two stages: Stage 1 adapts to over hundreds of public object assets, drawn either from the deployment scene alone or from all 101 scenes; Stage 2 personalizes to the user's own objects on the deployment scene's receptacles. Our strongest personalized policy reaches 62.5\% aggregate pickup success in the target scene. Deployed on a real Stretch 3, the same policy achieves 63.9\% success on a 12-task subset, closely tracking its 66.7\% simulated success on those tasks. We find that \shorthand{} is able to adapt to the target scene and objects from only simulation oracle data, suggesting an alternative way for personalizing VLA manipulation training using an automated real-to-sim-to-real approach.

\end{abstract}

\section{INTRODUCTION}

Consider a mobile manipulator deployed in a user's home. The user wants it to fetch a pill bottle from the shelf and a can of soup from the side table. Generalist VLAs promise this kind of open-world capability~\cite{intelligence_$_0.5$:_2025, ye2026worldactionmodelszeroshot}, but deployment exposes a gap: in our experiments, pretrained $\pi_{0.5}$ achieves 0.0\% pickup success rate (SR) in two target scenes (Sec. \ref{result}). Closing this gap through fine-tuning is standard practice, but the required data has traditionally come from teleoperation, policy rollouts, or human intervention in the target environment~\cite{DBLP:journals/corr/abs-1810-02890, intelligence2025pi06vlalearnsexperience}, a cost that recurs for every new user, receptacle, object, or configuration.

\begin{figure}[t]
\centering
\includegraphics[width=\columnwidth]{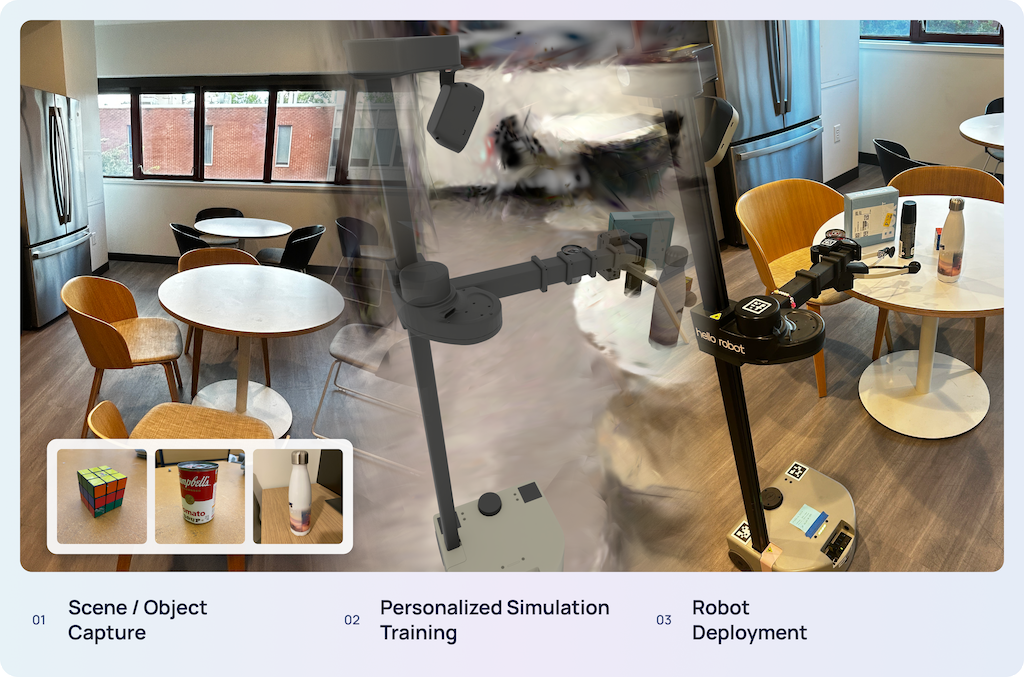}
\caption{An overview of \shorthand{}. From a handheld scan of a deployment scene and images of user-selected objects, \shorthand{} builds a simulation-ready digital twin, generates pickup demonstrations without any teleoperation, and fine-tunes a VLA for deployment in the captured scene.}
\label{fig:teaser}
\end{figure}

Simulation demonstrations and evaluation metrics help further improve generalist models, but they often rely on mesh-based rendering of generic scenes with simplified lighting conditions, matching deployment configurations marginally~\cite{nasiriany_robocasa:_2024, feng_libero:_2023}. Current real-to-sim-to-real systems use scene reconstructions through 3D Gaussian Splatting (3DGS) to improve visual fidelity and close the sim-to-real gap through demonstration generation and evaluation with privileged information~\cite{kim2026legs, jain_polaris:_2025, sato2026dreamdeploymenttimedemonstrationgeneration, jiang_gsworld:_2025}. However, these either still require human demonstration or annotation, or are often reconstructed at the level of individual tabletops, making it still laborious for deployment in a user's full environment. 

To address these challenges, we present \textbf{\shorthand{}}, an automated real-to-sim-to-real pipeline for VLA manipulation training and evaluation within a large building-scale scene and custom objects. The pipeline combines 3D Gaussian Splatting (3DGS) appearances with collision geometry from scenes scanned with a hand-held device. Target and distractor objects can be either chosen from publicly available 3D assets or generated using single images of real objects. The pipeline detects valid receptacles and uses a privileged-state geometric oracle to generate successful pickup demonstrations in simulation. The resulting environments also provide task resets and success predicates for policy evaluation. The only human input to \shorthand{} is a 20-30 minute handheld scan of the scene plus one image and the dimensions of each target object; no robot, operator expertise, or per-task repetition is needed. The same half hour of teleoperation would yield demonstrations for only one receptacle and one object set, before accounting for robot setup and operator training.

Our pipeline scales to 101 scanned indoor environments with a median of 104.87 m$^2$ navigable area, from which we generate 68,021 successful pickup demonstrations across 852 automatically discovered receptacles. We will make this dataset publicly available upon acceptance. Using a selected subset of this data, we study different approaches to personalization across broad adaptation sources and conduct sim-to-real experiments in target scenes and out-of-distribution (OOD) environments. We find that same-scene broad adaptation followed by personalization yields the strongest policy, improving over direct personalization by 17.5 percentage points on aggregate and reaching 63.9\% real-world SR where it scores 66.7\% in simulation, and a personalized policy fine-tuned from diverse multi-scene adaptation retains robustness to unseen receptacles

Our contributions are:
\begin{itemize}
\item An automated real-to-sim-to-real pipeline that converts scene captures and real objects into feasible manipulation tasks and demonstrations with no human effort beyond scene and object scans.
\item A large-scale simulation dataset generated from our pipeline, constructed from 101 building-scale indoor scenes with 852 diverse receptacles, with corresponding task metadata and demonstrations.
\item An empirical study on how broad simulation adaptation and personalization affect pickup performance on a user's scene and objects, with controlled simulation and real-world evaluation.
\end{itemize}

\section{Related Work}

\subsection{Vision-Language-Action Models and Adaptation}
Vision-language-action (VLA) models combine pretrained visual and linguistic representations with robot demonstrations to support transferable manipulation policies. OpenVLA provides an open model that can be fine-tuned for new tasks and embodiments~\cite{kim_openvla:_2024}; $\pi_{0}$ introduces a VLM-backed flow-matching architecture for continuous control, and $\pi_{0.5}$ uses heterogeneous co-training to improve generalization to unfamiliar environments~\cite{black_$_0$:_2024, intelligence_$_0.5$:_2025}. Even with broad heterogeneous pretraining, downstream adaptation remains important for policy improvement. OpenVLA-OFT studies action representations, parallel decoding, action chunking, and training objectives that improve fine-tuning efficiency and performance~\cite{kim_fine-tuning_2025}. Personalization also has a perceptual dimension, as Visual Attentive Prompting grounds user-specific objects from reference images and guides frozen VLAs through visual highlighting and instruction rewriting~\cite{lee2026bringcuppersonalizingvisionlanguageaction}. Our setting couples personalized object assets with adaptation to the appearance, receptacles, and feasible pickup configurations of a captured deployment scene.

\subsection{Automated and Deployment-Specific Data Generation}
Large real-world datasets such as Open X-Embodiment and DROID provide diverse manipulation experience, but collecting demonstrations across robots and environments requires substantial human effort~\cite{oneill_open_2024, khazatsky_droid:_2024}. MimicGen and DemoGen amplify limited human demonstrations by adapting trajectories and observations to new spatial configurations~\cite{mandlekar_mimicgen:_2023, xue_demogen:_2025}. Generative simulation methods address a different source of scale: GenSim generates task programs and expert demonstrations, RoboGen generates tasks, scenes, and supervision, and RoboTwin 2.0 combines automated execution-code generation with structured domain randomization for bimanual manipulation~\cite{wang_gensim:_2023, wang_robogen:_2023, chen2025robotwin}. These approaches establish the value of automated data generation, but deployment personalization additionally requires grounding the generated experience in the environment where the policy will operate.

\subsection{Real-to-Sim-to-Real Manipulation}
Reconstructed environments support both policy improvement and controlled evaluation. RialTo uses scanned digital twins, inverse distillation of real demonstrations, and simulation RL to robustify manipulation policies~\cite{torne_reconciling_2024}. EmbodiedSplat personalizes navigation policies in deployment environments reconstructed from mobile-device captures~\cite{chhablani_embodiedsplat:_2025}; our work builds on this paradigm for receptacle-local manipulation. For evaluation, SIMPLER studies how reducing visual and control discrepancies makes simulated policy performance predictive of real-world behavior, while PolaRiS constructs interactive environments from real-world captures for scalable policy evaluation~\cite{li_evaluating_2024, jain_polaris:_2025}.

Photorealistic rendering has emerged as an enabling component of real-to-sim-to-real manipulation. Particularly, 3D Gaussian Splatting (3DGS) supports efficient novel view synthesis~\cite{kerbl_3d_2023}, and SplatSim, RE$^3$SIM, RoboSimGS, and GSWorld use Gaussian-based scene representations for simulated manipulation data or closed-loop policy interaction~\cite{qureshi_splatsim:_2024, han_re$^3$sim:_2025, zhao_high-fidelity_2025, jiang_gsworld:_2025}. RL-GSBridge also demonstrates real-to-sim-to-real visual RL with Gaussian splat rendering~\cite{wu_rl-gsbridge:_2024}.

Some recent works employ real-to-sim for synthetic data generation. PRISM constructs semantically and geometrically aligned digital cousin scenes from a user-scene image and a language instruction, then uses task-and-motion planning to synthesize executable demonstrations without human teleoperation~\cite{ko_prism:_2026}. DREAM reconstructs a captured workspace, translates a supplied instruction into symbolic goals and success criteria, and generates and augments planned trajectories for VLA fine-tuning without a task-specific human demonstration~\cite{sato2026dreamdeploymenttimedemonstrationgeneration}. SimFoundry reconstructs simulation-ready digital twins and generates object, scene, and task cousins, including feasible tasks proposed from the reconstructed objects and affordances~\cite{ranawaka_simfoundry:_2026}. 

Recent agentic real-to-sim systems further automate simulation construction and downstream interaction. Agentic Real2Sim uses VLM-directed reconstruction and simulator-in-the-loop refinement to reproduce observed robot interactions as physically executable simulation episodes~\cite{chen2026agenticreal2simphysicsbasedworld}. Real2Gym reconstructs interactive simulation environments from demonstrations, enabling agents to practice and acquire reusable manipulation skills~\cite{ren2026real2gymbuildinggymsvideos}. More closely related to our pipeline, ARSTAG reconstructs task-specific simulation scenes from an image and language instruction, then generates demonstrations with task-consistent augmentations for VLA adaptation~\cite{li2026arstagagenticreal2sim2realtaskspecific}. Unlike these pipelines that condition on tasks and demonstrations, Splat2Act autonomously discovers manipulation opportunities across building-scale captured environments and generates simulation data for deployment-scene personalization.

A closely related concurrent work, LEGS, combines a captured 3DGS background with separately reconstructed robot and task-object meshes, procedurally generates demonstrations from privileged object poses and task-specific motion primitives, and fine-tunes VLAs for humanoid loco-manipulation~\cite{kim2026legs}. While its demonstration generation is teleoperation-free, LEGS assumes task-relevant objects and placement targets are already instantiated in the simulator and specified by the scene configuration, rather than discovering interaction opportunities or receptacles from a larger reconstructed environment. Similarly, VLK synthesizes navigation and object-interaction trajectories in reconstructed indoor scenes and trains a policy to predict whole-body kinematics for execution by a humanoid tracker~\cite{wang_vlk:_2026}. VLK requires manually annotated semantic 3D bounding boxes and walkable regions to identify task-relevant objects, support surfaces, and feasible navigation space. 

Two differences separate \shorthand{} from these works. First, scene scale: PRISM, DREAM, RE$^3$SIM, and SplatSim reconstruct a single workspace or tabletop, while LEGS and VLK operate in room-scale scenes but require task objects, placement targets, or support surfaces to be specified or annotated in advance. \shorthand{} processes building-scale captures with a median of 104.87 m$^2$ navigable area and discovers receptacles from geometry alone, yielding a median of 7 usable receptacles per scene, including non-tabletop surfaces. Second, personalization as the objective: while recent real-to-sim pipelines increasingly automate task-specific scene construction, demonstration generation, and domain randomization, \shorthand{} focuses on adapting a generalist VLA to the user's captured environment and object inventory. It autonomously discovers interaction opportunities across building-scale scenes and studies how scene-specific and multi-scene simulation data affect personalization and transfer, for which we provide a full analysis in Section \ref{section5}. Table~\ref{tab:related_work} shows a comparison with relevant prior works.

\begin{table}[t]
\centering
\caption{Comparison with capture-conditioned manipulation pipelines}
\label{tab:related_work}
\scriptsize
\setlength{\tabcolsep}{0.8pt}
\renewcommand{\arraystretch}{1.16}

\newcommand{\rwhead}[1]{%
  \makebox[0pt][l]{\hspace{-7pt}\rotatebox{65}{#1}}%
}

\resizebox{\columnwidth}{!}{%
\begin{tabular}{@{}
  >{\raggedright\arraybackslash}p{62pt}
  >{\centering\arraybackslash}m{29pt}
  *{8}{>{\centering\arraybackslash}m{20pt}}
  @{}}
\hline
\textbf{Feature}
 & \rwhead{\textbf{Splat2Act}}
 & \rwhead{VLK~\cite{wang_vlk:_2026}}
 & \rwhead{LEGS~\cite{kim2026legs}}
 & \rwhead{DREAM~\cite{sato2026dreamdeploymenttimedemonstrationgeneration}}
 & \rwhead{PRISM~\cite{ko_prism:_2026}}
 & \rwhead{SimFoundry~\cite{ranawaka_simfoundry:_2026}}
 & \rwhead{RE$^3$SIM~\cite{han_re$^3$sim:_2025}}
 & \rwhead{RialTo~\cite{torne_reconciling_2024}}
 & \rwhead{ARSTAG~\cite{li2026arstagagenticreal2sim2realtaskspecific}} \\
\hline

Scene scale
 & \shortstack{\textbf{Building-}\\\textbf{scale}}
 & Room
 & Room
 & \shortstack{Table-\\top}
 & \shortstack{Work-\\space}
 & \shortstack{Table-\\top}
 & \shortstack{Table-\\top}
 & \shortstack{Work-\\space}
 & \shortstack{Table-\\top} \\
\hline

Fully automated after capture$^*$
 & \cmark & \xmark & \xmark & \xmark
 & \xmark & \xmark & \xmark & \xmark
 & \xmark \\

Scene-wide receptacle discovery
 & \cmark & \xmark & \xmark & \xmark
 & \xmark & \xmark & \xmark & \xmark
 & \xmark \\

Feasibility-aware task assembly$^\dagger$
 & \cmark & \cmark & \xmark & \xmark
 & \cmark & \cmark & \cmark & \xmark
 & \cmark \\

Teleop-free demonstrations
 & \cmark & \cmark & \cmark & \cmark
 & \cmark & \xmark & \cmark & \xmark
 & \cmark \\
\hline

User-object reconstruction
 & \cmark & \xmark & \cmark & \cmark
 & \xmark & \cmark & \cmark & \cmark
 & \cmark \\

Target-scene policy training
 & \cmark & \cmark & \cmark & \cmark
 & \cmark & \cmark & \cmark & \cmark
 & \cmark \\
\hline

100+ captured scenes
 & \cmark & \xmark & \xmark & \xmark
 & \xmark & \xmark & \xmark & \xmark
 & \xmark \\

Real-robot validation
 & \cmark & \cmark & \cmark & \cmark
 & \cmark & \cmark & \cmark & \cmark
 & \cmark \\
\hline
\end{tabular}%
}

\par\vspace{4pt}
\begin{minipage}{\columnwidth}
\scriptsize
\cmark: demonstrated; \xmark: not demonstrated under
the stated definition.
Workspace denotes a local fixed-arm work area.
$^*$Requires no task-specific instructions, demonstrations,
or manual annotations beyond scene/object capture.
$^\dagger$Uses geometric or robot constraints to select
valid initial placements before trajectory generation.
\end{minipage}

\end{table}


\begin{figure*}[t]
\centering
\includegraphics[width=2\columnwidth]{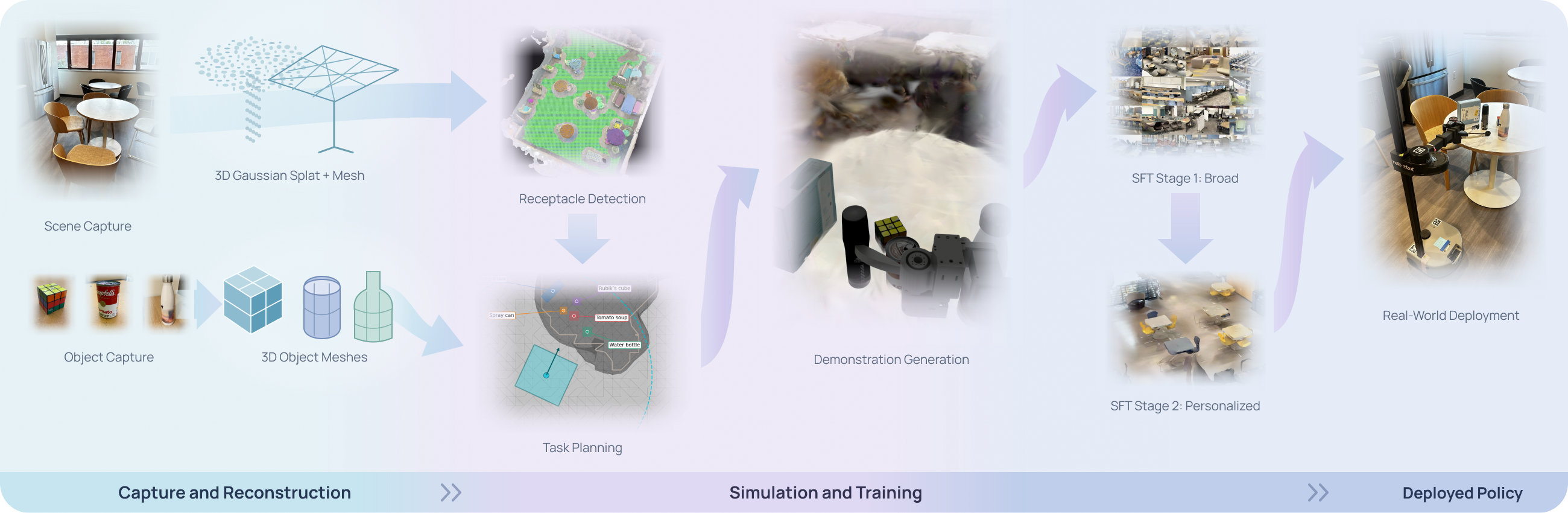}
\caption{The end-to-end pipeline of \shorthand{}. Scene captures are reconstructed as a 3DGS with a mesh collision geometry, while simulation-ready 3D assets are generated from object captures. Receptacle discovery and reachability-aware task planning construct feasible pickup tasks, and a privileged oracle generates demonstrations for broad Stage 1 adaptation and Stage 2 personalization before real-world deployment.}

\label{fig:pipeline}
\end{figure*}

\section{Method}

We study receptacle-local pickup with a mobile manipulator (Hello Robot Stretch 3~\cite{kemp_design_2022}). From a scene capture via a video on a standard phone and a separately chosen object inventory, we reconstruct the scene, autonomously extract receptacles, and construct executable tasks. Given the task, the oracle generates demonstrations for policy adaptation and personalization to the deployment scene. Only scene and object capture is required from a human. Fig.~\ref{fig:pipeline} shows the end-to-end pipeline of \shorthand{}.

\subsection{Scene and Object Representations}
Following EmbodiedSplat, we process the captured RGB-D sequence from an iPhone using Polycam to train a 3DGS representation for photorealistic rendering~\cite{kerbl_3d_2023}. For an average scene of our size, a capture usually takes 20-30 minutes. Because large-scene reconstruction requires significant geometric consistency, we use DN-Splatter for depth and normal regularization~\cite{turkulainen_dn-splatter:_2025}. We then extract a mesh from the trained Gaussian representation using SuGaR, a surface-aligned Gaussian meshing method~\cite{guedon_sugar:_2024}. This yields a hybrid scene representation: the Gaussian scene provides photorealistic egocentric visual observations, while the extracted mesh provides geometry for collision, receptacle discovery, and navigation analysis.

For object assets, we use two sources. Our broad adaptation stage filters 556 Objaverse assets compatible with the robot's workspace and gripper \cite{deitke_objaverse:_2023}. The personalization stage uses SAM3D to generate assets from single images of independently captured real objects~\cite{sam3dteam2025sam3d3dfyimages}. The meshes are then scripted to scale to user-specified dimensions. These objects are supplied separately from the scene capture and inserted as movable assets to provide object-specific appearance and shape without assuming exact reconstruction.

\subsection{Receptacle Discovery and Task Assembly}
Our receptacle detector searches over scene geometry for different heights and locations. Given the extracted mesh, we estimate the floor plane and retain upward-facing mesh regions whose heights and surface areas are compatible with placeable receptacles, determined by global thresholds based on the robot geometry. We cluster these candidate surface elements via HDBSCAN using their 3D positions and normals~\cite{hutchison_density-based_2013}, then merge and filter clusters according to geometric extent, surface stability, and height.

For each retained receptacle, we sample candidate object placement points over the surface region. We then compute a navigation mesh using the robot parameters and discard placement points that are not reachable from the robot’s navigable region, which is derived as the navmesh computed from Habitat Simulator~\cite{savva_habitat:_2019, szot_habitat_2021}. A candidate stand point is retained when at least 6 placement samples lie within the robot's horizontal reach radius of $0.70m$. We further test object settlement, grasp filtering, and oracle demonstration execution feasibility. We assemble configurations by varying robot pose, target pose, and distractors within the available regions.

\subsection{Demonstration Generation}
Our scripted oracle uses privileged object poses, geometry, and robot state. It transforms cached object-local antipodal grasp candidates into the live settled pose, filters candidates, and ranks feasible wrist and base configurations. A greedy planner selects local translation and yaw actions from a discrete action grid. The controller then servos arm extension, lift, and wrist yaw through pregrasp, descent, gripper closure, and lifting. If an attempt fails, it restores the attempt snapshot and retries another candidate. Given a scene, object assets, and task blueprints, the generation system autonomously initializes batched Isaac Sim environments~\cite{nvidia_isaac_2025}, plans and retries motions, evaluates outcomes, refills environments, and records successful demonstrations.
Completed slots receive new tasks while other slots continue. Physics runs at 100 Hz and control and camera recording at 20 Hz. 
At 64 parallel environments, the oracle is able to generate on average 60 successful pickup demonstrations per A40 GPU-hour, with variation from scene startup and exhaustive candidate retry.

\subsection{Policy Adaptation}

To adapt the policy to the task, environment, and robot embodiment, we use two-stage supervised fine-tuning (SFT). Stage 1 provides broad adaptation over hundreds of public object assets and automatically discovered receptacles, drawn either from the deployment scene alone or from all reconstructed scenes. Stage 2 personalizes the Stage 1 policy to the deployment scene using the user's own scanned objects on that scene's receptacles. Direct Stage-2 SFT from the pretrained $\pi_{0.5}$ serves as a comparison.

We fine-tune the $\pi_{0.5}$ policy with its conditional flow matching objective from successful oracle demonstrations~\cite{intelligence_$_0.5$:_2025, black_$_0$:_2024, lipman_flow_2022}. Given a ground truth action chunk $\mathbf{a} \in \mathbb{R}^{H \times d_a}$, we sample Gaussian noise $\boldsymbol{\epsilon} \sim \mathcal{N}(\mathbf{0}, I)$
and interpolation time $t \in [0,1]$, and form $\mathbf{x}_t = t\mathbf{a} + (1-t)\boldsymbol{\epsilon}$. The policy is trained using the supervised flow-matching loss $\mathcal{L}_{\mathrm{FM}} = \mathbb{E}_{\mathbf{a},\,\boldsymbol{\epsilon},\,t} \left[\left\|\mathbf{v}_\theta(\mathbf{x}_t, o, t) - (\mathbf{a} - \boldsymbol{\epsilon})\right\|_2^2\right]$. This trains the action expert to map noisy action chunks toward oracle action chunks.

We fully fine-tune the PaliGemma language backbone and the action expert while keeping the SigLIP image encoder frozen to preserve the pretrained visual-semantic prior, while allowing the action model to specialize to the reconstructed scene, robot embodiment, receptacle distribution, and task family~\cite{beyer_paligemma:_2024, zhai_sigmoid_2023}.

\subsection{Real-World Deployment}
In deployment scenarios, the robot streams observations to a remote policy server running on a single A40 GPU, and the server returns action chunks over a bidirectional gRPC interface. The deployment observation includes RGB images from head and wrist cameras, proprioceptive state, and a user prompt. The returned action chunk is executed synchronously, following prior works~\cite{black_$_0$:_2024, kim_openvla:_2024, brohan_rt-2:_2023, intelligence_$_0.5$:_2025, kim_fine-tuning_2025, ghosh_octo:_2024}, using position and velocity control.

\section{Datasets and Tasks}

\looseness=-1 We evaluate on two captured deployment scenes, Office and Conference, and use the remaining 99 scenes only as broad multi-scene adaptation data. Every task is instantiated under an easy condition (target alone on receptacle) and a hard condition (target plus distractors drawn from remaining object inventory); we report success pooled over both.
\subsection{Scene and Demonstrations}

Our scanned building-scale indoor scene collection contains 101 diverse reconstructions from a university campus with real lounges, kitchens, libraries, etc. We identified and filtered 852 final training receptacles in the retained broad training subset that comprises all 101 scenes. The scenes sum to 11,909.78 m$^2$ of navigable XY center-region area, with a median of 104.87 m$^2$ and 7 receptacles per scene.

Table~\ref{tab:source_data} reports the source generation outcomes. The two broad campaigns produce 68,021 successful pickup trajectories, totaling 273.46 recorded control hours. Across 852 discovered receptacles and 556 compatible object assets, the collection supports up to 473,712 candidate object-receptacle pairs before feasibility filtering. Each feasible pair can be further augmented through variations in object placement, robot initial pose, and distractor configurations. For illustration, generating 24 configurations per pair would yield approximately 11.4 million candidate tasks, corresponding to roughly 45,700 hours of demonstration data at our observed average trajectory duration, assuming all configurations were feasible and successful. In practice, we uniformly subsampled the multi-scene generation campaign by approximately 100$\times$ due to storage constraints, and the 68,021 collected demonstrations therefore represent only a fraction of this potential task space.

Our personalized task sources include the two deployment scenes, and seven real objects: pill bottle, Spam, tomato soup, cardboard box, Rubik's cube, spray can, and water bottle. These assets are spawned across seven receptacles in the Office scene and six receptacles in the Conference scene.

\begin{table}[t]
\centering
\caption{Oracle demonstration generation success rates}
\label{tab:source_data}
\small
\setlength{\tabcolsep}{3pt}
\begin{tabular}{lrrr}
\hline
Source & Successful & Assigned & Oracle SR \\
\hline
Broad Office & 32,052 & 40,478 & 79.18\% \\
Broad multi-scene & 35,969 & 55,067 & 65.32\% \\
Personalized Office & 3,781 & 4,086 & 92.54\% \\
Personalized Conference & 3,469 & 4,097 & 84.67\% \\
\hline
\end{tabular}
\end{table}

\subsection{Broad and Personalized Datasets}
\looseness=-1 For broad Stage 1 adaptation, we sample two subsets of 1,024 demonstrations each, one from Office and one from all 101 scenes, referred to as Office1k and Multi1k. Office1k spans 550 Objaverse target identities and seven supports in Office. Multi1k spans 556 identities, 101 scenes, and 852 receptacles. Our personalization stage uses three of the seven objects: pill bottle, Spam, and tomato soup. For any target object, the other six unused objects are eligible to be used as distractors (nearby objects to probe language steering capability).

Crossing the three target objects with seven designated Office receptacles gives 21 object-receptacle pairs. Similarly, Conference provides 18 object-receptacle pairs. Both scene datasets cover every designated pair. The Stage 2 personalization data separately contains 504 Office demonstrations and 500 Conference demonstrations, with each pair in Office augmented with 24 variations and around 28 augmentations per Conference pair. 

\subsection{Evaluation}

Personalized evaluation measures pickup of known target identities on designated receptacles under reassembled robot-target configurations disjoint from training. The designed task set contains one easy and one hard configuration for every object-receptacle pair. We further retain configurations with a successful state verified by our demonstration generation module. The task banks contain 40 Office (21 easy, 19 hard) and 35 Conference (18 easy, 17 hard) configurations, retaining all 39 object-receptacle pairs in total.

\looseness=-1 We evaluate each included checkpoint for 700 steps. The pickup predicate checks target elevation relative to its starting pose; success is evaluated at policy stop or the horizon and recorded explicitly before reset. A termination emitted by an unsuccessful policy counts as failure. We aggregate successful episodes divided by evaluated episodes across easy and hard conditions, preserving valid failures in the denominator.

\section{Experiments}
\label{section5}

Our experiments are designed to answer the following questions:
\begin{itemize}
    \item How do broad adaptation choices affect personalized pickup performance?
    \item How important is scene personalization?
    \item Do policies personalized in simulation transfer to the real deployment scene?
    \item How well do personalized policies generalize to a similar unseen scene?
\end{itemize}

\subsection{Experimental Setup}

All policies are fine-tuned from the same pretrained $\pi_{0.5}$ checkpoint, which consumes head and wrist RGB, discretized 4-dimensional proprioceptive state, and the task string, to predict chunks of 10 actions in a 10-dimensional Stretch action space with quantile-normalized actions and states. The action vector is
\(\mathbf{a}=[\Delta q_{\rm arm},0,\Delta q_{\rm lift},0,0,
\Delta\theta_{\rm wrist},g,\tau,\Delta x_{\rm base},
\Delta\theta_{\rm base}]\),
where \(g\) and \(\tau\) are the gripper and termination commands.

We use the same trajectory count for each stage of our training, but the resulting optimizer steps differ across conditions since trajectory length varies: Stage~1 Office1k and Multi1k train for 39{,}489 and 45{,}705 steps, and Stage~2 Office and Conference train for 15{,}963 and 17{,}390 steps. Every run trains for exactly one epoch at batch size 64 under a cosine schedule from $5\times10^{-5}$ to $5\times10^{-6}$.  All policies are queried at 20\,Hz and execute the full predicted chunk.

\subsection{Scaling Receptacles, Objects, and Augmentations}
\label{sec:scaling_exp}
We ablate the scale of Stage 2 data along three axes: the number of receptacles, rendered 3D objects used for training, and augmented tasks generated per object. Table~\ref{tab:scaling_receptacles} fixes four rendered 3D objects and varies the number of receptacles ($1$, $2$, $4$) and tasks per object ($8$, $16$, $24$) from Office1k. In-distribution evaluation uses the first two augmented tasks for each object-receptacle pair. Scaling receptacles and objects inevitably increases the number of distinct interaction configurations and, consequently, the amount of training data required to cover them. At only 8 tasks per object, success remains low across settings ($21.9$--$43.8\%$), whereas increasing augmentation depth consistently improves performance. With 24 tasks per object, success reaches $75.0\%$, $87.5\%$, and $75.0\%$ for one, two, and four receptacles, respectively, despite the corresponding corpus growing from 96 to 384 trajectories. These results suggest that trajectory scaling is a necessary consequence of personalizing a VLA over increasingly diverse receptacles and rendered objects, and that approximately 24 augmented trajectories per object-receptacle pair provides sufficient coverage to maintain strong in-distribution success as the personalization space grows. This ablation motivated the 24+ augmentations per pair used in Stage 2.

\begin{table}[t]
\centering
\caption{In-distribution pickup success as receptacle count and augmentation depth scale, with four rendered 3D objects used for training.}
\label{tab:scaling_receptacles}
\small
\setlength{\tabcolsep}{5pt}
\begin{tabular}{rrrr}
\hline
Rec. & Tasks/obj. & Trajs & Success Rate \\
\hline
1 & 8  & 32  & 25.0\% \\
1 & 16 & 64  & 75.0\% \\
1 & 24 & 96  & 75.0\% \\
2 & 8  & 64  & 43.8\% \\
2 & 16 & 128 & 75.0\% \\
2 & 24 & 192 & 87.5\% \\
4 & 8  & 128 & 21.9\% \\
4 & 16 & 256 & 59.4\% \\
4 & 24 & 384 & 75.0\% \\
\hline
\end{tabular}
\end{table}

\subsection{Broad Adaptation Sources}
\label{result}

\begin{figure}[t]
\centering
\includegraphics[width=\columnwidth]{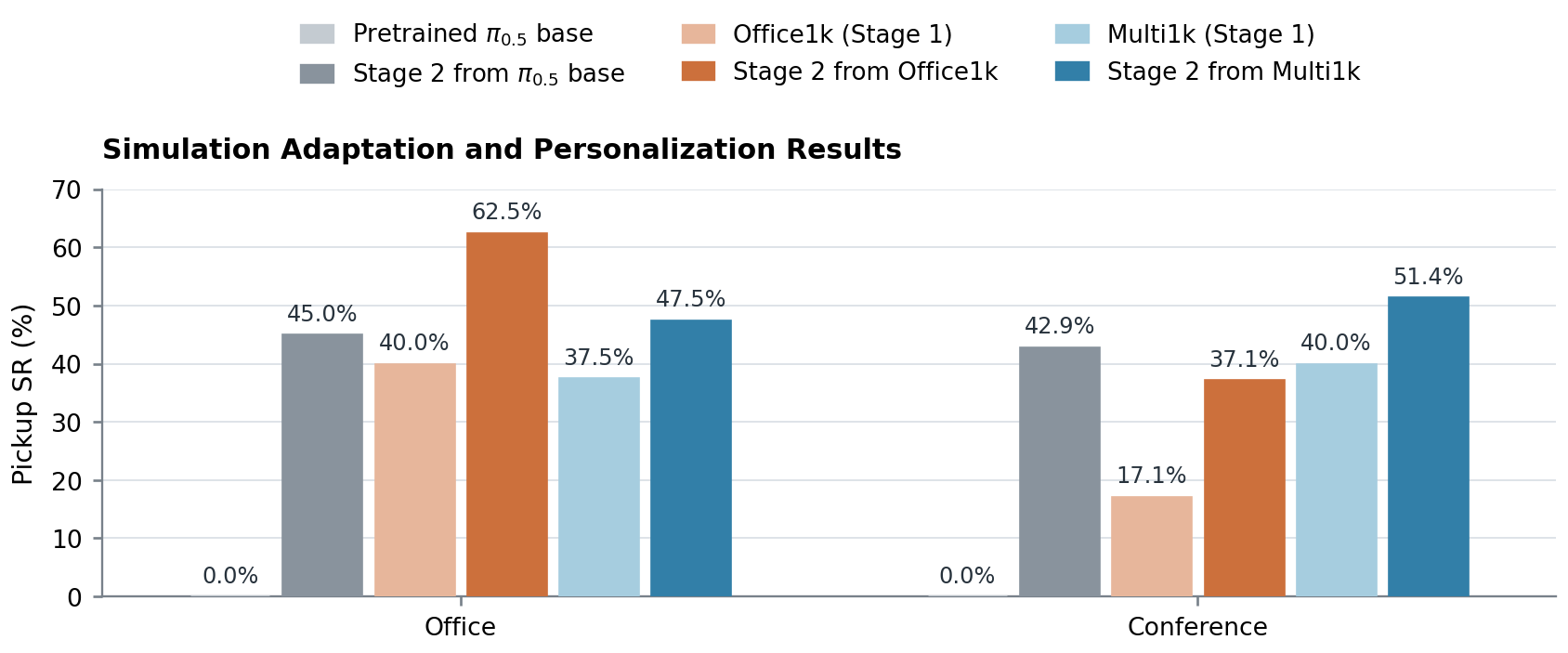}
\caption{Simulation evaluation on the personalized task banks. Office and Conference each show pretrained $\pi_{0.5}$ base, Office1k Stage 1, Multi1k Stage 1, and Stage 2 policies fine-tuned on them. Lighter bars show base/Stage 1 policies; darker bars show scene-matched personalized policies. Labels show SR (\%).}
\label{fig:adaptation_source}
\end{figure}

Our highest aggregate success rates are 62.5\% on the personalized Office Stage 2 policy fine-tuned from Office1k Stage 1 and 51.4\% on Conference Stage 2 policy fine-tuned from Multi1k Stage 1. Fig.~\ref{fig:adaptation_source} brings together pretrained base, Stage 1, and Stage 2 policies. To assess how broad adaptation affects personalized pickup, we also include the comparison among the three starting points: pretrained $\pi_{0.5}$ base, Office1k, and Multi1k. On Office, Stage 2 policies fine-tuned from Multi1k Stage 1 and $\pi_{0.5}$ base achieve 47.5\% and 45.0\%, respectively. On Conference, Stage 2 policies fine-tuned from $\pi_{0.5}$ base and Office1k Stage 1 achieve 42.9\% and 37.1\%.

\begin{figure}[t]
\centering
\includegraphics[width=\columnwidth]{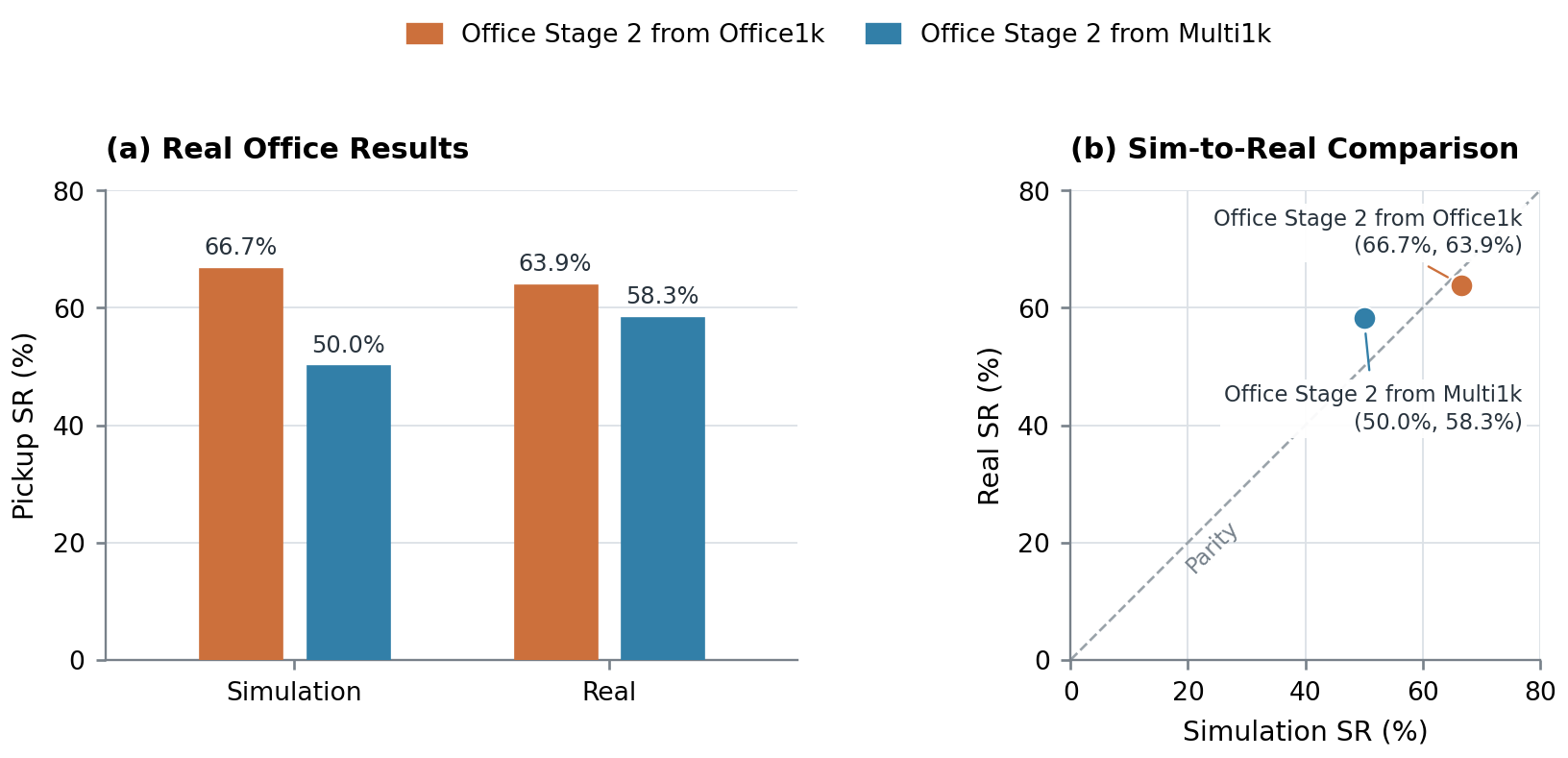}
\caption{Office sim-to-real comparison over 12 tasks. (a) Aggregate simulation and real SR. Each policy has 12 simulated episodes and 36 real attempts over the same object-receptacle-difficulty cells. (b) Policy-level scatter of those aggregate SRs; each point is one policy, and the dashed line indicates equal simulation and real SR. Labels show SR (\%).}
\label{fig:physical_office}
\end{figure}

The benefit of adding the broad adaptation stage depends on its data source and the deployment scene. The highest observed SR in each scene follows broad adaptation: on Office, Stage 2 from Office1k improves over Stage 2 from the $\pi_{0.5}$ base by 17.5 percentage points, and on Conference, Stage 2 from Multi1k improves over it by 8.6 percentage points. This suggests that same-scene Stage 1 training prepares the policy better than broad adaptation with diverse scenes and objects. However, broad adaptation does not uniformly help: the Stage 2 policy trained from Office1k Stage 1 is 5.7 percentage points below $\pi_{0.5}$ base Stage 2 on Conference. The easy/hard split (Table~\ref{tab:easy_hard}) shows where the broad stage helps most. Distractors cost every Stage 2 policy 10-26 percentage points in SR, but the penalty is largest for direct SFT from $\pi_{0.5}$ base and smallest after broad adaptation, so Stage 1 improves language grounding under clutter and not only pickup. The strongest policy reaches 71.4\% on easy and 52.6\% on hard Office tasks, reinforcing the superiority of same-scene adaptation and personalization.

The comparisons among Stage 1 policies, including $\pi_{0.5}$ base, in Fig.~\ref{fig:adaptation_source} provide context before personalization. The pretrained $\pi_{0.5}$ base achieves 0.0\% on both banks due to no exposure to the Stretch embodiment or our action space, so off-the-shelf deployment is not an option. Before personalization, Office1k Stage 1 achieves 40.0\% SR on Office but only 17.1\% on Conference, whereas Multi1k Stage 1 achieves 37.5\% and 40.0\%, respectively. This confirms that same-scene broad adaptation is the strongest starting point for its own scene, while multi-scene adaptation transfers better across scenes because of its diverse object-receptacle pairs.

\begin{table}[t]
\centering
\caption{Stage 2 pickup success by difficulty. Row headers give the number of evaluated configurations per scene.}
\label{tab:easy_hard}
\small
\setlength{\tabcolsep}{4pt}
\begin{tabular}{llrrr}
\hline
Scene & Stage 2 from & Easy & Hard & Easy$-$Hard \\
\hline
Office ($n$=21/19) & $\pi_{0.5}$ base & 57.1 & 31.6 & 25.6 \\
 & Office1k & \textbf{71.4} & \textbf{52.6} & 18.8 \\
 & Multi1k & 52.4 & 42.1 & 10.3 \\
\hline
Conference ($n$=18/17) & $\pi_{0.5}$ base & 55.6 & 29.4 & 26.1 \\
 & Office1k & 44.4 & 29.4 & 15.0 \\
 & Multi1k & \textbf{61.1} & \textbf{41.2} & 19.9 \\
\hline
\end{tabular}
\end{table}

\subsection{Scene Personalization}

\begin{figure}[ht]
\centering
\includegraphics[width=.85\columnwidth]{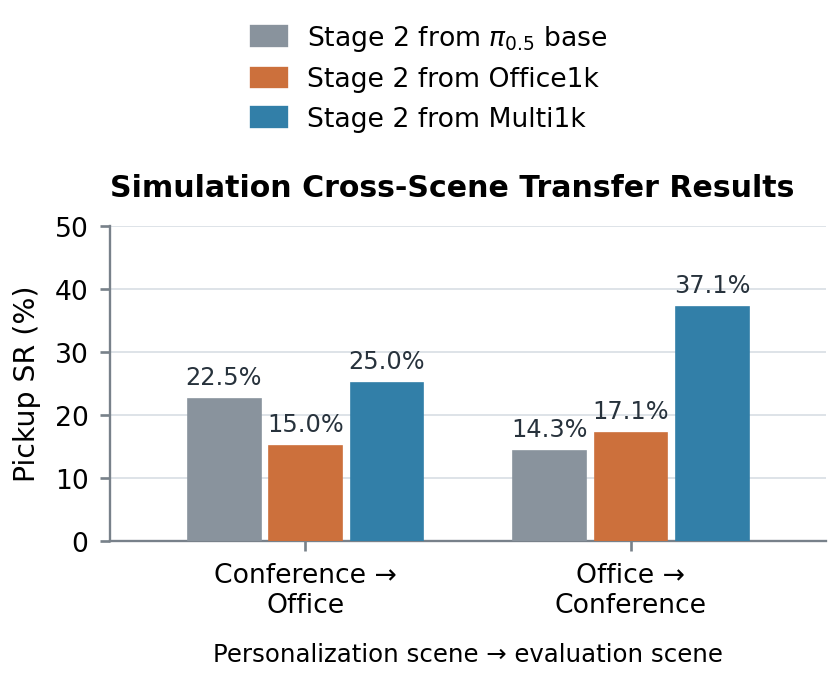}
\caption{Simulation cross-scene personalized-policy performance. The horizontal labels specify personalization scene $\rightarrow$ evaluation scene; each group compares Stage 2 policies fine-tuned from $\pi_{0.5}$ base, Office1k Stage 1 policy, and Multi1k Stage 1 policy. Labels show SR (\%).}
\label{fig:scene_personalization}
\end{figure}

Fig.~\ref{fig:scene_personalization} shows personalized Stage 2 policies performing on the other scene. On Office, Conference-personalized policies on $\pi_{0.5}$ base, Office1k Stage 1, and Multi1k Stage 1 achieve 22.5\%, 15.0\%, and 25.0\% SR, respectively. In the reverse direction, Office-personalized policies achieve 14.3\%, 17.1\%, and 37.1\% on Conference.

Comparing these scores with the scene-matched policies in Fig.~\ref{fig:adaptation_source}, matching the scene improves SR in all six comparisons by 14.3\%-47.5\%. For example, Office1k Stage 1 followed by Office personalization achieves 62.5\% SR on Office, compared with 15.0\% after Conference personalization. Personalizing to the wrong scene is worse than not personalizing at all: Office1k Stage 1 alone reaches 40.0\% on Office, but after Conference personalization it drops to 15.0\%. Object-only demonstrations without scene adaptation therefore actively harm deployment performance.

\subsection{Sim-to-Real Transfer in Office}

We evaluate the policies personalized from Multi1k and from Office1k on Stretch 3 over 12 tasks from 3 receptacles. We document three attempts per task for 36 total trials per policy in the real-world evaluation, and 1 attempt per task for 12 total trials per policy in the simulated evaluation. Successful trials require both target pickup and policy termination. Across the tasks, the Stage 2 policy fine-tuned from Office1k Stage 1 achieves 66.7\% SR in simulation and 63.9\% on the real robot, while the Stage 2 policy fine-tuned from Multi1k Stage 1 achieves 50.0\% and 58.3\%, respectively (Fig.~\ref{fig:physical_office}a).

\looseness=-1 The sim-to-real scatter (Fig.~\ref{fig:physical_office}b) shows the same two policy aggregates. Both policies transfer within 8 percentage points of their simulated SR and preserve their ordering, indicating that the reconstructed scene is faithful enough for policies fine-tuned by our pipeline to transfer beyond simulation. Qualitatively, the real robot shows less precise arm extension and grasp placement than in simulation, but real-world contact is more forgiving: the gripper's mechanical compliance allows it to deform during contact, and its rubber fingertips provide more contact friction than the simulated contact model, so imperfect grasping can still succeed.

\subsection{Generalization to the unseen}

We test the Office-personalized policies in an unseen lab on a black and a yellow table, using the same three targets, both easy and hard conditions on each receptacle, with three attempts per configuration. The black table is qualitatively closer to the Office receptacles in height and appearance, whereas the yellow table has a larger shift.

\begin{figure}[ht]
\centering
\includegraphics[width=.8\columnwidth]{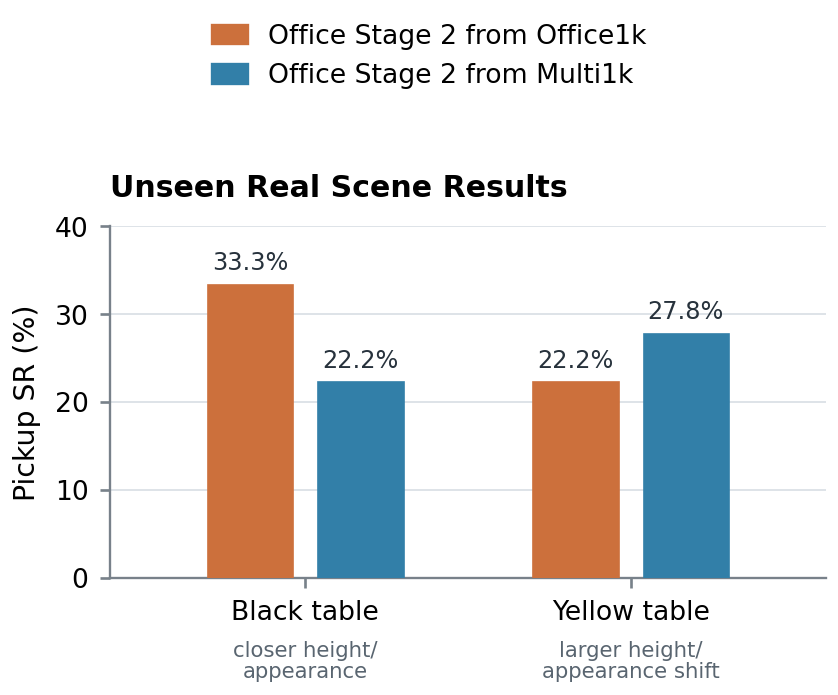}
\caption{Transfer to the unseen lab. Black table and Yellow table each include three attempts for each of three targets under both difficulties (18 trials per policy). Labels show SR (\%).}
\label{fig:unseen_receptacles}
\end{figure}

Fig.~\ref{fig:unseen_receptacles} shows the corresponding results averaged between easy and hard trials. Stage 2 Office policies trained on Office1k Stage 1 and Multi1k Stage 1 achieve 33.3\% and 22.2\% SR on the black table, and 22.2\% and 27.8\% on the yellow table, respectively. 

The ordering reverses between the two tables, so these results do not establish a consistent winner. The outcome shows that the personalized policy from scene-specific broad adaptation is more brittle in OOD scenarios, whereas the Stage 2 policy on Multi1k Stage 1 remains consistent across receptacles, showing its diverse Stage 1 adaptation maintains some robustness to receptacle shift. Relative to their 62.5\% and 47.5\% success in the personalization scene, both policies drop substantially, which reinforces that scene-specific personalization, not object familiarity alone, drives competent deployment performance.

\section{CONCLUSIONS}

In this work, we presented \shorthand{}, an automated real-to-sim-to-real manipulation pipeline for personalizing VLA policies to objects and scenes with no human input beyond scene and object capture. We demonstrated that our pipeline is capable of handling building-scale indoor scenes and generating tasks with demonstrations at scale. We further showed that personalizing a policy broadly adapted on the same scene with diverse objects can lead to robust behaviors in simulation and real-world deployment scenarios. We provided a comprehensive analysis of adaptation sources, sim-to-real transfer, and generalization ability. In the future, this pipeline can be applied to longer-horizon mobile manipulation tasks, with expectations for reinforcement learning to improve success rate further through exploration.








\bibliographystyle{IEEEtran}
\bibliography{references}

@IEEEtranBSTCTL{BSTcontrol,
  CTLuse_forced_etal = {yes},
  CTLmax_names_forced_etal = {6},
  CTLnames_show_etal = {1},
}

@misc{brohan_rt-2:_2023,
	title = {{RT}-2: {Vision}-{Language}-{Action} {Models} {Transfer} {Web} {Knowledge} to {Robotic} {Control}},
	copyright = {arXiv.org perpetual, non-exclusive license},
	shorttitle = {{RT}-2},
	note = {arXiv:2307.15818},
	doi = {10.48550/ARXIV.2307.15818},
	urldate = {2026-05-28},
	publisher = {arXiv},
	author = {Brohan, Anthony and Brown, Noah and Carbajal, Justice and Chebotar, Yevgen and Chen, Xi and Choromanski, Krzysztof and Ding, Tianli and Driess, Danny and Dubey, Avinava and Finn, Chelsea and Florence, Pete and Fu, Chuyuan and Arenas, Montse Gonzalez and Gopalakrishnan, Keerthana and Han, Kehang and Hausman, Karol and Herzog, Alexander and Hsu, Jasmine and Ichter, Brian and Irpan, Alex and Joshi, Nikhil and Julian, Ryan and Kalashnikov, Dmitry and Kuang, Yuheng and Leal, Isabel and Lee, Lisa and Lee, Tsang-Wei Edward and Levine, Sergey and Lu, Yao and Michalewski, Henryk and Mordatch, Igor and Pertsch, Karl and Rao, Kanishka and Reymann, Krista and Ryoo, Michael and Salazar, Grecia and Sanketi, Pannag and Sermanet, Pierre and Singh, Jaspiar and Singh, Anikait and Soricut, Radu and Tran, Huong and Vanhoucke, Vincent and Vuong, Quan and Wahid, Ayzaan and Welker, Stefan and Wohlhart, Paul and Wu, Jialin and Xia, Fei and Xiao, Ted and Xu, Peng and Xu, Sichun and Yu, Tianhe and Zitkovich, Brianna},
	year = {2023},
}

@misc{li2026arstagagenticreal2sim2realtaskspecific,
      title={ARSTAG: An Agentic Real2Sim2Real System for Task-Specific Robot Data Generation}, 
      author={Bowei Li and Yuner Zhang and Changliu Liu},
      year={2026},
      eprint={2609.24563},
      archivePrefix={arXiv},
      primaryClass={cs.RO},
      url={https://arxiv.org/abs/2609.24563}, 
}

@misc{chen2026agenticreal2simphysicsbasedworld,
      title={Agentic Real2Sim: Physics-based World Modeling with Vision-Language Agents}, 
      author={Guanxiong Chen and Qianjun Xia and Jiawei Peng and Heng Zhang and Pengyu Jing and Bole Ma and Justin Qian and Yixian Cheng and Ziyi Jiao and Bingyang Zhou and Yiduo Qu and Luoxin Ye and Kaifeng Zhang and Kunyi Wang and Weijia Zeng and Yunuo Chen and Pengzhi Yang and Ziqiu Zeng and Siyuan Luo and Huamin Wang and Chao Liu and Alan Yuille and Fan Shi and Changxi Zheng and Yunzhu Li and Chenfanfu Jiang and Peter Yichen Chen},
      year={2026},
      eprint={2607.19190},
      archivePrefix={arXiv},
      primaryClass={cs.RO},
      url={https://arxiv.org/abs/2607.19190}, 
}

@misc{ren2026real2gymbuildinggymsvideos,
      title={Real2Gym: Building Gyms from Videos, Bringing Skills to Robots}, 
      author={Kerui Ren and Yingxiang Xu and Kaiwen Song and Linning Xu and Bo Dai and Mulin Yu and Tao Lu},
      year={2026},
      eprint={2609.37089},
      archivePrefix={arXiv},
      primaryClass={cs.CV},
      url={https://arxiv.org/abs/2609.37089}, 
}

@misc{kim_openvla:_2024,
	title = {{OpenVLA}: {An} {Open}-{Source} {Vision}-{Language}-{Action} {Model}},
	copyright = {Creative Commons Attribution 4.0 International},
	shorttitle = {{OpenVLA}},
	note = {arXiv:2406.09246},
	doi = {10.48550/ARXIV.2406.09246},
	urldate = {2026-05-28},
	publisher = {arXiv},
	author = {Kim, Moo Jin and Pertsch, Karl and Karamcheti, Siddharth and Xiao, Ted and Balakrishna, Ashwin and Nair, Suraj and Rafailov, Rafael and Foster, Ethan and Lam, Grace and Sanketi, Pannag and Vuong, Quan and Kollar, Thomas and Burchfiel, Benjamin and Tedrake, Russ and Sadigh, Dorsa and Levine, Sergey and Liang, Percy and Finn, Chelsea},
	year = {2024},
}

@inproceedings{ghosh_octo:_2024,
	title = {Octo: {An} {Open}-{Source} {Generalist} {Robot} {Policy}},
	isbn = {9798990284807},
	shorttitle = {Octo},
	url = {http://www.roboticsproceedings.org/rss20/p090.pdf},
	doi = {10.15607/RSS.2024.XX.090},
	urldate = {2026-05-28},
	booktitle = {Robotics: {Science} and {Systems} {XX}},
	publisher = {Robotics: Science and Systems Foundation},
	author = {Ghosh, Dibya and Walke, Homer and Pertsch, Karl and Black, Kevin and Mees, Oier and Dasari, Sudeep and Hejna, Joey and Kreiman, Tobias and Xu, Charles and Luo, Jianlan and Tan, You and Chen, Lawrence and Vuong, Quan and Xiao, Ted and Sanketi, Pannag and Sadigh, Dorsa and Finn, Chelsea and Levine, Sergey},
	month = jul,
	year = {2024},
}

@misc{black_$_0$:_2024,
	title = {{$\pi_0$}: {A} {Vision}-{Language}-{Action} {Flow} {Model} for {General} {Robot} {Control}},
	copyright = {arXiv.org perpetual, non-exclusive license},
	shorttitle = {{$\pi_0$}},
	note = {arXiv:2410.24164},
	doi = {10.48550/ARXIV.2410.24164},
	urldate = {2026-05-28},
	publisher = {arXiv},
	author = {Black, Kevin and Brown, Noah and Driess, Danny and Esmail, Adnan and Equi, Michael and Finn, Chelsea and Fusai, Niccolo and Groom, Lachy and Hausman, Karol and Ichter, Brian and Jakubczak, Szymon and Jones, Tim and Ke, Liyiming and Levine, Sergey and Li-Bell, Adrian and Mothukuri, Mohith and Nair, Suraj and Pertsch, Karl and Shi, Lucy Xiaoyang and Tanner, James and Vuong, Quan and Walling, Anna and Wang, Haohuan and Zhilinsky, Ury},
	year = {2024},
}

@misc{intelligence_$_0.5$:_2025,
	title = {{$\pi_{0.5}$}: a {Vision}-{Language}-{Action} {Model} with {Open}-{World} {Generalization}},
	copyright = {arXiv.org perpetual, non-exclusive license},
	shorttitle = {{$\pi_{0.5}$}},
    note   = {arXiv:2504.16054},
	doi = {10.48550/ARXIV.2504.16054},
	urldate = {2026-05-28},
	publisher = {arXiv},
	author = {{Physical Intelligence} and Black, Kevin and Brown, Noah and Darpinian, James and Dhabalia, Karan and Driess, Danny and Esmail, Adnan and Equi, Michael and Finn, Chelsea and Fusai, Niccolo and Galliker, Manuel Y. and Ghosh, Dibya and Groom, Lachy and Hausman, Karol and Ichter, Brian and Jakubczak, Szymon and Jones, Tim and Ke, Liyiming and LeBlanc, Devin and Levine, Sergey and Li-Bell, Adrian and Mothukuri, Mohith and Nair, Suraj and Pertsch, Karl and Ren, Allen Z. and Shi, Lucy Xiaoyang and Smith, Laura and Springenberg, Jost Tobias and Stachowicz, Kyle and Tanner, James and Vuong, Quan and Walke, Homer and Walling, Anna and Wang, Haohuan and Yu, Lili and Zhilinsky, Ury},
	year = {2025},
}

@inproceedings{kim_fine-tuning_2025,
	title = {Fine-{Tuning} {Vision}-{Language}-{Action} {Models}: {Optimizing} {Speed} and {Success}},
	isbn = {9798990284814},
	shorttitle = {Fine-{Tuning} {Vision}-{Language}-{Action} {Models}},
	url = {https://www.roboticsproceedings.org/rss21/p017.pdf},
	doi = {10.15607/RSS.2025.XXI.017},
	urldate = {2026-05-28},
	booktitle = {Robotics: {Science} and {Systems} {XXI}},
	publisher = {Robotics: Science and Systems Foundation},
	author = {Kim, Moo and Finn, Chelsea and Liang, Percy},
	month = jun,
	year = {2025},
}

@inproceedings{oneill_open_2024,
	address = {Yokohama, Japan},
	title = {Open {X}-{Embodiment}: {Robotic} {Learning} {Datasets} and {RT}-{X} {Models}},
	isbn = {9798350384574},
	shorttitle = {Open {X}-{Embodiment}},
	url = {https://ieeexplore.ieee.org/document/10611477/},
	doi = {10.1109/ICRA57147.2024.10611477},
	urldate = {2026-09-17},
	booktitle = {2024 {IEEE} {International} {Conference} on {Robotics} and {Automation} ({ICRA})},
	publisher = {IEEE},
	author = {O’Neill, Abby and Rehman, Abdul and Maddukuri, Abhiram and Gupta, Abhishek and Padalkar, Abhishek and Lee, Abraham and Pooley, Acorn and Gupta, Agrim and Mandlekar, Ajay and Jain, Ajinkya and Tung, Albert and Bewley, Alex and Herzog, Alex and Irpan, Alex and Khazatsky, Alexander and Rai, Anant and Gupta, Anchit and Wang, Andrew and Singh, Anikait and Garg, Animesh and Kembhavi, Aniruddha and Xie, Annie and Brohan, Anthony and Raffin, Antonin and Sharma, Archit and Yavary, Arefeh and Jain, Arhan and Balakrishna, Ashwin and Wahid, Ayzaan and Burgess-Limerick, Ben and Kim, Beomjoon and Schölkopf, Bernhard and Wulfe, Blake and Ichter, Brian and Lu, Cewu and Xu, Charles and Le, Charlotte and Finn, Chelsea and Wang, Chen and Xu, Chenfeng and Chi, Cheng and Huang, Chenguang and Chan, Christine and Agia, Christopher and Pan, Chuer and Fu, Chuyuan and Devin, Coline and Xu, Danfei and Morton, Daniel and Driess, Danny and Chen, Daphne and Pathak, Deepak and Shah, Dhruv and Büchler, Dieter and Jayaraman, Dinesh and Kalashnikov, Dmitry and Sadigh, Dorsa and Johns, Edward and Foster, Ethan and Liu, Fangchen and Ceola, Federico and Xia, Fei and Zhao, Feiyu and Stulp, Freek and Zhou, Gaoyue and Sukhatme, Gaurav S. and Salhotra, Gautam and Yan, Ge and Feng, Gilbert and Schiavi, Giulio and Berseth, Glen and Kahn, Gregory and Wang, Guanzhi and Su, Hao and Fang, Hao-Shu and Shi, Haochen and Bao, Henghui and Ben Amor, Heni and Christensen, Henrik I and Furuta, Hiroki and Walke, Homer and Fang, Hongjie and Ha, Huy and Mordatch, Igor and Radosavovic, Ilija and Leal, Isabel and Liang, Jacky and Abou-Chakra, Jad and Kim, Jaehyung and Drake, Jaimyn and Peters, Jan and Schneider, Jan and Hsu, Jasmine and Bohg, Jeannette and Bingham, Jeffrey and Wu, Jeffrey and Gao, Jensen and Hu, Jiaheng and Wu, Jiajun and Wu, Jialin and Sun, Jiankai and Luo, Jianlan and Gu, Jiayuan and Tan, Jie and Oh, Jihoon and Wu, Jimmy and Lu, Jingpei and Yang, Jingyun and Malik, Jitendra and Silvério, João and Hejna, Joey and Booher, Jonathan and Tompson, Jonathan and Yang, Jonathan and Salvador, Jordi and Lim, Joseph J. and Han, Junhyek and Wang, Kaiyuan and Rao, Kanishka and Pertsch, Karl and Hausman, Karol and Go, Keegan and Gopalakrishnan, Keerthana and Goldberg, Ken and Byrne, Kendra and Oslund, Kenneth and Kawaharazuka, Kento and Black, Kevin and Lin, Kevin and Zhang, Kevin and Ehsani, Kiana and Lekkala, Kiran and Ellis, Kirsty and Rana, Krishan and Srinivasan, Krishnan and Fang, Kuan and Singh, Kunal Pratap and Zeng, Kuo-Hao and Hatch, Kyle and Hsu, Kyle and Itti, Laurent and Chen, Lawrence Yunliang and Pinto, Lerrel and Fei-Fei, Li and Tan, Liam and Fan, Linxi Jim and Ott, Lionel and Lee, Lisa and Weihs, Luca and Chen, Magnum and Lepert, Marion and Memmel, Marius and Tomizuka, Masayoshi and Itkina, Masha and Castro, Mateo Guaman and Spero, Max and Du, Maximilian and Ahn, Michael and Yip, Michael C. and Zhang, Mingtong and Ding, Mingyu and Heo, Minho and Srirama, Mohan Kumar and Sharma, Mohit and Kim, Moo Jin and Kanazawa, Naoaki and Hansen, Nicklas and Heess, Nicolas and Joshi, Nikhil J and Suenderhauf, Niko and Liu, Ning and Di Palo, Norman and Shafiullah, Nur Muhammad Mahi and Mees, Oier and Kroemer, Oliver and Bastani, Osbert and Sanketi, Pannag R and Miller, Patrick Tree and Yin, Patrick and Wohlhart, Paul and Xu, Peng and Fagan, Peter David and Mitrano, Peter and Sermanet, Pierre and Abbeel, Pieter and Sundaresan, Priya and Chen, Qiuyu and Vuong, Quan and Rafailov, Rafael and Tian, Ran and Doshi, Ria and Martín-Martín, Roberto and Baijal, Rohan and Scalise, Rosario and Hendrix, Rose and Lin, Roy and Qian, Runjia and Zhang, Ruohan and Mendonca, Russell and Shah, Rutav and Hoque, Ryan and Julian, Ryan and Bustamante, Samuel and Kirmani, Sean and Levine, Sergey and Lin, Shan and Moore, Sherry and Bahl, Shikhar and Dass, Shivin and Sonawani, Shubham and Song, Shuran and Xu, Sichun and Haldar, Siddhant and Karamcheti, Siddharth and Adebola, Simeon and Guist, Simon and Nasiriany, Soroush and Schaal, Stefan and Welker, Stefan and Tian, Stephen and Ramamoorthy, Subramanian and Dasari, Sudeep and Belkhale, Suneel and Park, Sungjae and Nair, Suraj and Mirchandani, Suvir and Osa, Takayuki and Gupta, Tanmay and Harada, Tatsuya and Matsushima, Tatsuya and Xiao, Ted and Kollar, Thomas and Yu, Tianhe and Ding, Tianli and Davchev, Todor and Zhao, Tony Z. and Armstrong, Travis and Darrell, Trevor and Chung, Trinity and Jain, Vidhi and Vanhoucke, Vincent and Zhan, Wei and Zhou, Wenxuan and Burgard, Wolfram and Chen, Xi and Wang, Xiaolong and Zhu, Xinghao and Geng, Xinyang and Liu, Xiyuan and Liangwei, Xu and Li, Xuanlin and Lu, Yao and Ma, Yecheng Jason and Kim, Yejin and Chebotar, Yevgen and Zhou, Yifan and Zhu, Yifeng and Wu, Yilin and Xu, Ying and Wang, Yixuan and Bisk, Yonatan and Cho, Yoonyoung and Lee, Youngwoon and Cui, Yuchen and Cao, Yue and Wu, Yueh-Hua and Tang, Yujin and Zhu, Yuke and Zhang, Yunchu and Jiang, Yunfan and Li, Yunshuang and Li, Yunzhu and Iwasawa, Yusuke and Matsuo, Yutaka and Ma, Zehan and Xu, Zhuo and Cui, Zichen Jeff and Zhang, Zichen and Lin, Zipeng},
	month = may,
	year = {2024},
	pages = {6892--6903},
}

@inproceedings{khazatsky_droid:_2024,
	title = {{DROID}: {A} {Large}-{Scale} {In}-{The}-{Wild} {Robot} {Manipulation} {Dataset}},
	isbn = {9798990284807},
	shorttitle = {{DROID}},
	url = {http://www.roboticsproceedings.org/rss20/p120.pdf},
	doi = {10.15607/RSS.2024.XX.120},
	urldate = {2026-05-28},
	booktitle = {Robotics: {Science} and {Systems} {XX}},
	publisher = {Robotics: Science and Systems Foundation},
	author = {Khazatsky, Alexander and Pertsch, Karl and Nair, Suraj and Balakrishna, Ashwin and Dasari, Sudeep and Karamcheti, Siddharth and Nasiriany, Soroush and Srirama, Mohan and Chen, Lawrence and Ellis, Kirsty and Fagan, Peter and Hejna, Joey and Itkina, Masha and Lepert, Marion and Ma, Yecheng and Miller, Patrick and Wu, Jimmy and Belkhale, Suneel and Dass, Shivin and Ha, Huy and Jain, Arhan and Lee, Abraham and Lee, Youngwoon and Memmel, Marius and Park, Sungjae and Radosavovic, Ilija and Wang, Kaiyuan and Zhan, Albert and Black, Kevin and Chi, Cheng and Hatch, Kyle and Lin, Shan and Lu, Jingpei and Mercat, Jean and Rehman, Abdul and Sanketi, Pannag and Sharma, Archit and Simpson, Cody and Vuong, Quan and Walke, Homer and Wulfe, Blake and Xiao, Ted and Yang, Jonathan and Yavary, Arefeh and Zhao, Tony and Agia, Christopher and Baijal, Rohan and Castro, Mateo and Chen, Daphne and Chen, Qiuyu and Chung, Trinity and Drake, Jaimyn and Foster, Ethan and Gao, Jensen and Herrera, David and Heo, Minho and Hsu, Kyle and Hu, Jiaheng and Jackson, Donovon and Le, Charlotte and Li, Yunshuang and Lin, Roy and Ma, Zehan and Maddukuri, Abhiram and Mirchandani, Suvir and Morton, Daniel and Nguyen, Tony and O'Neill, Abigail and Scalise, Rosario and Seale, Derick and Son, Victor and Tian, Stephen and Tran, Emi and Wang, Andrew and Wu, Yilin and Xie, Annie and Yang, Jingyun and Yin, Patrick and Zhang, Yunchu and Bastani, Osbert and Berseth, Glen and Bohg, Jeannette and Goldberg, Ken and Gupta, Abhinav and Gupta, Abhishek and Jayaraman, Dinesh and Lim, Joseph and Malik, Jitendra and Martín-Martín, Roberto and Ramamoorthy, Subramanian and Sadigh, Dorsa and Song, Shuran and Wu, Jiajun and Yip, Michael and Zhu, Yuke and Kollar, Thomas and Levine, Sergey and Finn, Chelsea},
	month = jul,
	year = {2024},
}

@misc{mandlekar_mimicgen:_2023,
	title = {{MimicGen}: {A} {Data} {Generation} {System} for {Scalable} {Robot} {Learning} using {Human} {Demonstrations}},
	copyright = {arXiv.org perpetual, non-exclusive license},
	shorttitle = {{MimicGen}},
	note = {arXiv:2310.17596},
	doi = {10.48550/ARXIV.2310.17596},
	urldate = {2026-05-28},
	publisher = {arXiv},
	author = {Mandlekar, Ajay and Nasiriany, Soroush and Wen, Bowen and Akinola, Iretiayo and Narang, Yashraj and Fan, Linxi and Zhu, Yuke and Fox, Dieter},
	year = {2023},
}

@inproceedings{nasiriany_robocasa:_2024,
	title = {{RoboCasa}: {Large}-{Scale} {Simulation} of {Household} {Tasks} for {Generalist} {Robots}},
	isbn = {9798990284807},
	shorttitle = {{RoboCasa}},
	url = {http://www.roboticsproceedings.org/rss20/p050.pdf},
	doi = {10.15607/RSS.2024.XX.050},
	urldate = {2026-05-28},
	booktitle = {Robotics: {Science} and {Systems} {XX}},
	publisher = {Robotics: Science and Systems Foundation},
	author = {Nasiriany, Soroush and Maddukuri, Abhiram and Zhang, Lance and Parikh, Adeet and Lo, Aaron and Joshi, Abhishek and Mandlekar, Ajay and Zhu, Yuke},
	month = jul,
	year = {2024},
}

@article{wang_gensim:_2023,
	title = {{GenSim}: {Generating} {Robotic} {Simulation} {Tasks} via {Large} {Language} {Models}},
	copyright = {arXiv.org perpetual, non-exclusive license},
	shorttitle = {{GenSim}},
	note = {arXiv:2310.01361},
	doi = {10.48550/ARXIV.2310.01361},
	urldate = {2026-05-28},
	author = {Wang, Lirui and Ling, Yiyang and Yuan, Zhecheng and Shridhar, Mohit and Bao, Chen and Qin, Yuzhe and Wang, Bailin and Xu, Huazhe and Wang, Xiaolong},
	year = {2023},
}

@misc{wang_robogen:_2023,
	title = {{RoboGen}: {Towards} {Unleashing} {Infinite} {Data} for {Automated} {Robot} {Learning} via {Generative} {Simulation}},
	copyright = {arXiv.org perpetual, non-exclusive license},
	shorttitle = {{RoboGen}},
	note = {arXiv:2311.01455},
	doi = {10.48550/ARXIV.2311.01455},
	urldate = {2026-05-28},
	publisher = {arXiv},
	author = {Wang, Yufei and Xian, Zhou and Chen, Feng and Wang, Tsun-Hsuan and Wang, Yian and Fragkiadaki, Katerina and Erickson, Zackory and Held, David and Gan, Chuang},
	year = {2023},
}

@inproceedings{xue_demogen:_2025,
	title = {{DemoGen}: {Synthetic} {Demonstration} {Generation} for {Data}-{Efficient} {Visuomotor} {Policy} {Learning}},
	isbn = {9798990284814},
	shorttitle = {{DemoGen}},
	url = {https://www.roboticsproceedings.org/rss21/p157.pdf},
	doi = {10.15607/RSS.2025.XXI.157},
	urldate = {2026-05-28},
	booktitle = {Robotics: {Science} and {Systems} {XXI}},
	publisher = {Robotics: Science and Systems Foundation},
	author = {Xue, Zhengrong and Deng, Shuying and Chen, Zhenyang and Wang, Yixuan and Yuan, Zhecheng and Xu, Huazhe},
	month = jun,
	year = {2025},
}

@misc{li_evaluating_2024,
	title = {Evaluating {Real}-{World} {Robot} {Manipulation} {Policies} in {Simulation}},
	copyright = {Creative Commons Attribution 4.0 International},
	note = {arXiv:2405.05941},
	doi = {10.48550/ARXIV.2405.05941},
	urldate = {2026-05-28},
	publisher = {arXiv},
	author = {Li, Xuanlin and Hsu, Kyle and Gu, Jiayuan and Pertsch, Karl and Mees, Oier and Walke, Homer Rich and Fu, Chuyuan and Lunawat, Ishikaa and Sieh, Isabel and Kirmani, Sean and Levine, Sergey and Wu, Jiajun and Finn, Chelsea and Su, Hao and Vuong, Quan and Xiao, Ted},
	year = {2024},
}

@misc{jain_polaris:_2025,
	title = {{PolaRiS}: {Scalable} {Real}-to-{Sim} {Evaluations} for {Generalist} {Robot} {Policies}},
	copyright = {Creative Commons Zero v1.0 Universal},
	shorttitle = {{PolaRiS}},
	note = {arXiv:2512.16881},
	doi = {10.48550/ARXIV.2512.16881},
	urldate = {2026-05-28},
	publisher = {arXiv},
	author = {Jain, Arhan and Zhang, Mingtong and Arora, Kanav and Chen, William and Torne, Marcel and Irshad, Muhammad Zubair and Zakharov, Sergey and Wang, Yue and Levine, Sergey and Finn, Chelsea and Ma, Wei-Chiu and Shah, Dhruv and Gupta, Abhishek and Pertsch, Karl},
	year = {2025},
}

@misc{chhablani_embodiedsplat:_2025,
	title = {{EmbodiedSplat}: {Personalized} {Real}-to-{Sim}-to-{Real} {Navigation} with {Gaussian} {Splats} from a {Mobile} {Device}},
	copyright = {Creative Commons Attribution Non Commercial Share Alike 4.0 International},
	shorttitle = {{EmbodiedSplat}},
	note = {arXiv:2509.17430},
	doi = {10.48550/ARXIV.2509.17430},
	urldate = {2026-05-28},
	publisher = {arXiv},
	author = {Chhablani, Gunjan and Ye, Xiaomeng and Irshad, Muhammad Zubair and Kira, Zsolt},
	year = {2025},
}

@misc{torne_reconciling_2024,
	title = {Reconciling {Reality} through {Simulation}: {A} {Real}-to-{Sim}-to-{Real} {Approach} for {Robust} {Manipulation}},
	copyright = {arXiv.org perpetual, non-exclusive license},
	shorttitle = {Reconciling {Reality} through {Simulation}},
	note = {arXiv:2403.03949},
	doi = {10.48550/ARXIV.2403.03949},
	urldate = {2026-05-28},
	publisher = {arXiv},
	author = {Torne, Marcel and Simeonov, Anthony and Li, Zechu and Chan, April and Chen, Tao and Gupta, Abhishek and Agrawal, Pulkit},
	year = {2024},
}

@misc{han_re$^3$sim:_2025,
	title = {{RE$^3$SIM}: {Generating} {High}-{Fidelity} {Simulation} {Data} via {3D}-{Photorealistic} {Real}-to-{Sim} for {Robotic} {Manipulation}},
	copyright = {Creative Commons Attribution Non Commercial No Derivatives 4.0 International},
	shorttitle = {{RE$^3$SIM}},
	note = {arXiv:2502.08645},
	doi = {10.48550/ARXIV.2502.08645},
	urldate = {2026-05-28},
	publisher = {arXiv},
	author = {Han, Xiaoshen and Liu, Minghuan and Chen, Yilun and Yu, Junqiu and Lyu, Xiaoyang and Tian, Yang and Wang, Bolun and Zhang, Weinan and Pang, Jiangmiao},
	year = {2025},
}

@misc{zhao_high-fidelity_2025,
	title = {High-{Fidelity} {Simulated} {Data} {Generation} for {Real}-{World} {Zero}-{Shot} {Robotic} {Manipulation} {Learning} with {Gaussian} {Splatting}},
	copyright = {arXiv.org perpetual, non-exclusive license},
	note = {arXiv:2510.10637},
	doi = {10.48550/ARXIV.2510.10637},
	urldate = {2026-05-28},
	publisher = {arXiv},
	author = {Zhao, Haoyu and Zeng, Cheng and Zhuang, Linghao and Zhao, Yaxi and Xue, Shengke and Wang, Hao and Zhao, Xingyue and Li, Zhongyu and Li, Kehan and Huang, Siteng and Chen, Mingxiu and Li, Xin and Zhao, Deli and Zou, Hua},
	year = {2025},
}

@article{kerbl_3d_2023,
	title = {{3D} {Gaussian} {Splatting} for {Real}-{Time} {Radiance} {Field} {Rendering}},
	volume = {42},
	issn = {0730-0301, 1557-7368},
	url = {https://dl.acm.org/doi/10.1145/3592433},
	doi = {10.1145/3592433},
	language = {en},
	number = {4},
	urldate = {2026-05-28},
	journal = {ACM Transactions on Graphics},
	author = {Kerbl, Bernhard and Kopanas, Georgios and Leimkuehler, Thomas and Drettakis, George},
	month = aug,
	year = {2023},
	pages = {1--14},
}

@inproceedings{turkulainen_dn-splatter:_2025,
	address = {Tucson, AZ, USA},
	title = {{DN}-{Splatter}: {Depth} and {Normal} {Priors} for {Gaussian} {Splatting} and {Meshing}},
	copyright = {https://doi.org/10.15223/policy-029},
	isbn = {9798331510831},
	shorttitle = {{DN}-{Splatter}},
	url = {https://ieeexplore.ieee.org/document/10943388/},
	doi = {10.1109/WACV61041.2025.00241},
	urldate = {2026-05-28},
	booktitle = {2025 {IEEE}/{CVF} {Winter} {Conference} on {Applications} of {Computer} {Vision} ({WACV})},
	publisher = {IEEE},
	author = {Turkulainen, Matias and Ren, Xuqian and Melekhov, Iaroslav and Seiskari, Otto and Rahtu, Esa and Kannala, Juho},
	month = feb,
	year = {2025},
	pages = {2421--2431},
}

@misc{qureshi_splatsim:_2024,
	title = {{SplatSim}: {Zero}-{Shot} {Sim2Real} {Transfer} of {RGB} {Manipulation} {Policies} {Using} {Gaussian} {Splatting}},
	copyright = {Creative Commons Attribution Non Commercial Share Alike 4.0 International},
	shorttitle = {{SplatSim}},
	note = {arXiv:2409.10161},
	doi = {10.48550/ARXIV.2409.10161},
	urldate = {2026-05-28},
	publisher = {arXiv},
	author = {Qureshi, Mohammad Nomaan and Garg, Sparsh and Yandun, Francisco and Held, David and Kantor, George and Silwal, Abhisesh},
	year = {2024},
}

@misc{wu_rl-gsbridge:_2024,
	title = {{RL}-{GSBridge}: {3D} {Gaussian} {Splatting} {Based} {Real2Sim2Real} {Method} for {Robotic} {Manipulation} {Learning}},
	copyright = {arXiv.org perpetual, non-exclusive license},
	shorttitle = {{RL}-{GSBridge}},
	note = {arXiv:2409.20291},
	doi = {10.48550/ARXIV.2409.20291},
	urldate = {2026-05-28},
	publisher = {arXiv},
	author = {Wu, Yuxuan and Pan, Lei and Wu, Wenhua and Wang, Guangming and Miao, Yanzi and Xu, Fan and Wang, Hesheng},
	year = {2024},
}

@misc{jiang_gsworld:_2025,
	title = {{GSWorld}: {Closed}-{Loop} {Photo}-{Realistic} {Simulation} {Suite} for {Robotic} {Manipulation}},
	copyright = {arXiv.org perpetual, non-exclusive license},
	shorttitle = {{GSWorld}},
	note = {arXiv:2510.20813},
	doi = {10.48550/ARXIV.2510.20813},
	urldate = {2026-05-28},
	publisher = {arXiv},
	author = {Jiang, Guangqi and Chang, Haoran and Qiu, Ri-Zhao and Liang, Yutong and Ji, Mazeyu and Zhu, Jiyue and Dong, Zhao and Zou, Xueyan and Wang, Xiaolong},
	year = {2025},
}

@inproceedings{kemp_design_2022,
	address = {Philadelphia, PA, USA},
	title = {The {Design} of {Stretch}: {A} {Compact}, {Lightweight} {Mobile} {Manipulator} for {Indoor} {Human} {Environments}},
	copyright = {https://doi.org/10.15223/policy-029},
	isbn = {9781728196817},
	shorttitle = {The {Design} of {Stretch}},
	url = {https://ieeexplore.ieee.org/document/9811922/},
	doi = {10.1109/ICRA46639.2022.9811922},
	urldate = {2026-05-28},
	booktitle = {2022 {International} {Conference} on {Robotics} and {Automation} ({ICRA})},
	publisher = {IEEE},
	author = {Kemp, Charles C. and Edsinger, Aaron and Clever, Henry M. and Matulevich, Blaine},
	month = may,
	year = {2022},
	pages = {3150--3157},
}

@inproceedings{guedon_sugar:_2024,
	address = {Seattle, WA, USA},
	title = {{SuGaR}: {Surface}-{Aligned} {Gaussian} {Splatting} for {Efficient} {3D} {Mesh} {Reconstruction} and {High}-{Quality} {Mesh} {Rendering}},
	copyright = {https://doi.org/10.15223/policy-029},
	isbn = {9798350353006},
	shorttitle = {{SuGaR}},
	url = {https://ieeexplore.ieee.org/document/10655755/},
	doi = {10.1109/CVPR52733.2024.00512},
	urldate = {2026-05-28},
	booktitle = {2024 {IEEE}/{CVF} {Conference} on {Computer} {Vision} and {Pattern} {Recognition} ({CVPR})},
	publisher = {IEEE},
	author = {Guédon, Antoine and Lepetit, Vincent},
	month = jun,
	year = {2024},
	pages = {5354--5363},
}

@inproceedings{savva_habitat:_2019,
	address = {Seoul, Korea (South)},
	title = {Habitat: {A} {Platform} for {Embodied} {AI} {Research}},
	copyright = {https://ieeexplore.ieee.org/Xplorehelp/downloads/license-information/IEEE.html},
	isbn = {9781728148038},
	shorttitle = {Habitat},
	url = {https://ieeexplore.ieee.org/document/9010745/},
	doi = {10.1109/ICCV.2019.00943},
	urldate = {2026-05-28},
	booktitle = {2019 {IEEE}/{CVF} {International} {Conference} on {Computer} {Vision} ({ICCV})},
	publisher = {IEEE},
	author = {Savva, Manolis and Kadian, Abhishek and Maksymets, Oleksandr and Zhao, Yili and Wijmans, Erik and Jain, Bhavana and Straub, Julian and Liu, Jia and Koltun, Vladlen and Malik, Jitendra and Parikh, Devi and Batra, Dhruv},
	month = oct,
	year = {2019},
	pages = {9338--9346},
}

@misc{szot_habitat_2021,
	title = {Habitat 2.0: {Training} {Home} {Assistants} to {Rearrange} their {Habitat}},
	copyright = {arXiv.org perpetual, non-exclusive license},
	shorttitle = {Habitat 2.0},
	note = {arXiv:2106.14405},
	doi = {10.48550/ARXIV.2106.14405},
	urldate = {2026-05-28},
	publisher = {arXiv},
	author = {Szot, Andrew and Clegg, Alex and Undersander, Eric and Wijmans, Erik and Zhao, Yili and Turner, John and Maestre, Noah and Mukadam, Mustafa and Chaplot, Devendra and Maksymets, Oleksandr and Gokaslan, Aaron and Vondrus, Vladimir and Dharur, Sameer and Meier, Franziska and Galuba, Wojciech and Chang, Angel and Kira, Zsolt and Koltun, Vladlen and Malik, Jitendra and Savva, Manolis and Batra, Dhruv},
	year = {2021},
}

@inproceedings{deitke_objaverse:_2023,
	address = {Vancouver, BC, Canada},
	title = {Objaverse: {A} {Universe} of {Annotated} {3D} {Objects}},
	copyright = {https://doi.org/10.15223/policy-029},
	isbn = {9798350301298},
	shorttitle = {Objaverse},
	url = {https://ieeexplore.ieee.org/document/10205395/},
	doi = {10.1109/CVPR52729.2023.01263},
	urldate = {2026-05-28},
	booktitle = {2023 {IEEE}/{CVF} {Conference} on {Computer} {Vision} and {Pattern} {Recognition} ({CVPR})},
	publisher = {IEEE},
	author = {Deitke, Matt and Schwenk, Dustin and Salvador, Jordi and Weihs, Luca and Michel, Oscar and VanderBilt, Eli and Schmidt, Ludwig and Ehsanit, Kiana and Kembhavi, Aniruddha and Farhadi, Ali},
	month = jun,
	year = {2023},
	pages = {13142--13153},
}

@inproceedings{feng_libero:_2023,
	address = {New Orleans, Louisiana, USA},
	title = {{LIBERO}: {Benchmarking} {Knowledge} {Transfer} for {Lifelong} {Robot} {Learning}},
	isbn = {9781713899112},
	shorttitle = {{LIBERO}},
	url = {http://www.proceedings.com/075280-1939.html},
	doi = {10.52202/075280-1939},
	urldate = {2026-05-28},
	booktitle = {Advances in {Neural} {Information} {Processing} {Systems} 36},
	publisher = {Neural Information Processing Systems Foundation, Inc. (NeurIPS)},
	author = {Feng, Yihao and Gao, Chongkai and Liu, Bo and Liu, Qiang and Stone, Peter and Zhu, Yifeng and Zhu, Yuke},
	year = {2023},
	pages = {44776--44791},
}

@misc{nvidia_isaac_2025,
	title = {Isaac {Lab}: {A} {GPU}-{Accelerated} {Simulation} {Framework} for {Multi}-{Modal} {Robot} {Learning}},
	copyright = {Creative Commons Attribution 4.0 International},
	shorttitle = {Isaac {Lab}},
	note = {arXiv:2511.04831},
	doi = {10.48550/ARXIV.2511.04831},
	urldate = {2026-05-28},
	publisher = {arXiv},
	author = {{NVIDIA} and Mittal, Mayank and Roth, Pascal and Tigue, James and Richard, Antoine and Zhang, Octi and Du, Peter and Serrano-Muñoz, Antonio and Yao, Xinjie and Zurbrügg, René and Rudin, Nikita and Wawrzyniak, Lukasz and Rakhsha, Milad and Denzler, Alain and Heiden, Eric and Borovicka, Ales and Ahmed, Ossama and Akinola, Iretiayo and Anwar, Abrar and Carlson, Mark T. and Feng, Ji Yuan and Garg, Animesh and Gasoto, Renato and Gulich, Lionel and Guo, Yijie and Gussert, M. and Hansen, Alex and Kulkarni, Mihir and Li, Chenran and Liu, Wei and Makoviychuk, Viktor and Malczyk, Grzegorz and Mazhar, Hammad and Moghani, Masoud and Murali, Adithyavairavan and Noseworthy, Michael and Poddubny, Alexander and Ratliff, Nathan and Rehberg, Welf and Schwarke, Clemens and Singh, Ritvik and Smith, James Latham and Tang, Bingjie and Thaker, Ruchik and Trepte, Matthew and Van Wyk, Karl and Yu, Fangzhou and Millane, Alex and Ramasamy, Vikram and Steiner, Remo and Subramanian, Sangeeta and Volk, Clemens and Chen, CY and Jawale, Neel and Kuruttukulam, Ashwin Varghese and Lin, Michael A. and Mandlekar, Ajay and Patzwaldt, Karsten and Welsh, John and Zhao, Huihua and Anes, Fatima and Lafleche, Jean-Francois and Moënne-Loccoz, Nicolas and Park, Soowan and Stepinski, Rob and Van Gelder, Dirk and Amevor, Chris and Carius, Jan and Chang, Jumyung and Chen, Anka He and Ciechomski, Pablo de Heras and Daviet, Gilles and Mohajerani, Mohammad and von Muralt, Julia and Reutskyy, Viktor and Sauter, Michael and Schirm, Simon and Shi, Eric L. and Terdiman, Pierre and Vilella, Kenny and Widmer, Tobias and Yeoman, Gordon and Chen, Tiffany and Grizan, Sergey and Li, Cathy and Li, Lotus and Smith, Connor and Wiltz, Rafael and Alexis, Kostas and Chang, Yan and Chu, David and Fan, Linxi "Jim" and Farshidian, Farbod and Handa, Ankur and Huang, Spencer and Hutter, Marco and Narang, Yashraj and Pouya, Soha and Sheng, Shiwei and Zhu, Yuke and Macklin, Miles and Moravanszky, Adam and Reist, Philipp and Guo, Yunrong and Hoeller, David and State, Gavriel},
	year = {2025},
}

@misc{lipman_flow_2022,
	title = {Flow {Matching} for {Generative} {Modeling}},
	copyright = {arXiv.org perpetual, non-exclusive license},
	note = {arXiv:2210.02747},
	doi = {10.48550/ARXIV.2210.02747},
	urldate = {2026-05-28},
	publisher = {arXiv},
	author = {Lipman, Yaron and Chen, Ricky T. Q. and Ben-Hamu, Heli and Nickel, Maximilian and Le, Matt},
	year = {2022},
}

@misc{beyer_paligemma:_2024,
	title = {{PaliGemma}: {A} versatile {3B} {VLM} for transfer},
	copyright = {Creative Commons Attribution 4.0 International},
	shorttitle = {{PaliGemma}},
	note = {arXiv:2407.07726},
	doi = {10.48550/ARXIV.2407.07726},
	urldate = {2026-05-28},
	publisher = {arXiv},
	author = {Beyer, Lucas and Steiner, Andreas and Pinto, André Susano and Kolesnikov, Alexander and Wang, Xiao and Salz, Daniel and Neumann, Maxim and Alabdulmohsin, Ibrahim and Tschannen, Michael and Bugliarello, Emanuele and Unterthiner, Thomas and Keysers, Daniel and Koppula, Skanda and Liu, Fangyu and Grycner, Adam and Gritsenko, Alexey and Houlsby, Neil and Kumar, Manoj and Rong, Keran and Eisenschlos, Julian and Kabra, Rishabh and Bauer, Matthias and Bošnjak, Matko and Chen, Xi and Minderer, Matthias and Voigtlaender, Paul and Bica, Ioana and Balazevic, Ivana and Puigcerver, Joan and Papalampidi, Pinelopi and Henaff, Olivier and Xiong, Xi and Soricut, Radu and Harmsen, Jeremiah and Zhai, Xiaohua},
	year = {2024},
}

@inproceedings{zhai_sigmoid_2023,
	address = {Paris, France},
	title = {Sigmoid {Loss} for {Language} {Image} {Pre}-{Training}},
	copyright = {https://doi.org/10.15223/policy-029},
	isbn = {9798350307184},
	url = {https://ieeexplore.ieee.org/document/10377550/},
	doi = {10.1109/ICCV51070.2023.01100},
	urldate = {2026-05-28},
	booktitle = {2023 {IEEE}/{CVF} {International} {Conference} on {Computer} {Vision} ({ICCV})},
	publisher = {IEEE},
	author = {Zhai, Xiaohua and Mustafa, Basil and Kolesnikov, Alexander and Beyer, Lucas},
	month = oct,
	year = {2023},
	pages = {11941--11952},
}

@incollection{hutchison_density-based_2013,
	address = {Berlin, Heidelberg},
	title = {Density-{Based} {Clustering} {Based} on {Hierarchical} {Density} {Estimates}},
	volume = {7819},
	isbn = {9783642374555 9783642374562},
	url = {http://link.springer.com/10.1007/978-3-642-37456-2_14},
	urldate = {2026-05-28},
	booktitle = {Advances in {Knowledge} {Discovery} and {Data} {Mining}},
	publisher = {Springer Berlin Heidelberg},
	author = {Campello, Ricardo J. G. B. and Moulavi, Davoud and Sander, Joerg},
    editor = {Pei, Jian and Tseng, Vincent S. and Cao, Longbing and Motoda, Hiroshi and Xu, Guandong},
	year = {2013},
	doi = {10.1007/978-3-642-37456-2_14},
	pages = {160--172},
}

@article{sam3dteam2025sam3d3dfyimages,
      title={SAM 3D: 3Dfy Anything in Images}, 
      author={{SAM 3D Team} and Xingyu Chen and Fu-Jen Chu and Pierre Gleize and Kevin J Liang and Alexander Sax and Hao Tang and Weiyao Wang and Michelle Guo and Thibaut Hardin and Xiang Li and Aohan Lin and Jiawei Liu and Ziqi Ma and Anushka Sagar and Bowen Song and Xiaodong Wang and Jianing Yang and Bowen Zhang and Piotr Dollár and Georgia Gkioxari and Matt Feiszli and Jitendra Malik},
      year={2025},
      eprint={2511.16624},
      archivePrefix={arXiv},
      primaryClass={cs.CV},
      note = {arXiv:2511.16624}, 
}

@misc{ye2026worldactionmodelszeroshot,
      title={World Action Models are Zero-shot Policies},
      author={Seonghyeon Ye and Yunhao Ge and Kaiyuan Zheng and Shenyuan Gao and Sihyun Yu and George Kurian and Suneel Indupuru and You Liang Tan and Chuning Zhu and Jiannan Xiang and Ayaan Malik and Kyungmin Lee and William Liang and Nadun Ranawaka and Jiasheng Gu and Yinzhen Xu and Guanzhi Wang and Fengyuan Hu and Avnish Narayan and Johan Bjorck and Jing Wang and Gwanghyun Kim and Dantong Niu and Ruijie Zheng and Yuqi Xie and Jimmy Wu and Qi Wang and Ryan Julian and Danfei Xu and Yilun Du and Yevgen Chebotar and Scott Reed and Jan Kautz and Yuke Zhu and Linxi "Jim" Fan and Joel Jang},
      year={2026},
      eprint={2602.15922},
      archivePrefix={arXiv},
      primaryClass={cs.RO},
      note = {arXiv:2602.15922},
}

@article{DBLP:journals/corr/abs-1810-02890,
  author       = {Michael Kelly and
                  Chelsea Sidrane and
                  Katherine Rose Driggs{-}Campbell and
                  Mykel J. Kochenderfer},
  title        = {HG-DAgger: Interactive Imitation Learning with Human Experts},
  journal      = {CoRR},
  volume       = {abs/1810.02890},
  year         = {2018},
  eprinttype   = {arXiv},
  eprint       = {1810.02890},
  bibsource    = {dblp computer science bibliography, https://dblp.org}
}

@misc{intelligence2025pi06vlalearnsexperience,
      title={$\pi^{*}_{0.6}$: a VLA That Learns From Experience}, 
      author={{Physical Intelligence} and Ali Amin and Raichelle Aniceto and Ashwin Balakrishna and Kevin Black and Ken Conley and Grace Connors and James Darpinian and Karan Dhabalia and Jared DiCarlo and Danny Driess and Michael Equi and Adnan Esmail and Yunhao Fang and Chelsea Finn and Catherine Glossop and Thomas Godden and Ivan Goryachev and Lachy Groom and Hunter Hancock and Karol Hausman and Gashon Hussein and Brian Ichter and Szymon Jakubczak and Rowan Jen and Tim Jones and Ben Katz and Liyiming Ke and Chandra Kuchi and Marinda Lamb and Devin LeBlanc and Sergey Levine and Adrian Li-Bell and Yao Lu and Vishnu Mano and Mohith Mothukuri and Suraj Nair and Karl Pertsch and Allen Z. Ren and Charvi Sharma and Lucy Xiaoyang Shi and Laura Smith and Jost Tobias Springenberg and Kyle Stachowicz and Will Stoeckle and Alex Swerdlow and James Tanner and Marcel Torne and Quan Vuong and Anna Walling and Haohuan Wang and Blake Williams and Sukwon Yoo and Lili Yu and Ury Zhilinsky and Zhiyuan Zhou},
      year={2025},
      eprint={2511.14759},
      archivePrefix={arXiv},
      primaryClass={cs.LG},
      note = {arXiv:2511.14759}, 
}

@article{kim2026legs,
  title   = {LEGS: Fine-Tuning Teleop-Free VLAs for Humanoid Loco-manipulation in an Embodied Gaussian Splatting World},
  author  = {Kim, Hojune and Chen, Timothy and Sun, Jiankai and Osterberg, Lars W. and Chen, Qianzhong and Wang, Ke and Schwager, Mac},
  journal = {arXiv preprint arXiv:2606.01458},
  year    = {2026}
}

@misc{sato2026dreamdeploymenttimedemonstrationgeneration,
      title={DREAM: Deployment-Time Demonstration Generation via Real-to-Sim for Scalable Policy Adaptation}, 
      author={Makoto Sato and Tatsuya Matsushima and Yutaka Matsuo and Yusuke Iwasawa},
      year={2026},
      eprint={2608.29078},
      archivePrefix={arXiv},
      primaryClass={cs.RO},
      note = {arXiv:2608.29078}, 
}

@misc{lee2026bringcuppersonalizingvisionlanguageaction,
      title={Bring My Cup! Personalizing Vision-Language-Action Models with Visual Attentive Prompting}, 
      author={Sangoh Lee and Sangwoo Mo and Wook-Shin Han},
      year={2026},
      eprint={2512.20014},
      archivePrefix={arXiv},
      primaryClass={cs.RO},
      note = {arXiv:2512.20014}, 
}

@article{chen2025robotwin,
  title={Robotwin 2.0: A scalable data generator and benchmark with strong domain randomization for robust bimanual robotic manipulation},
  author={Chen, Tianxing and Chen, Zanxin and Chen, Baijun and Cai, Zijian and Liu, Yibin and Li, Zixuan and Liang, Qiwei and Lin, Xianliang and Ge, Yiheng and Gu, Zhenyu and others},
  journal={arXiv preprint arXiv:2506.18088},
  year={2025}
}

@misc{ko_prism:_2026,
	title = {{PRISM}: {Personalized} {Robotic} {Dataset} {Generation} via {Image}-based {Scene} and {Motion} {Synthesis}},
	copyright = {Creative Commons Attribution 4.0 International},
	shorttitle = {{PRISM}},
	note = {arXiv:2607.04880},
	doi = {10.48550/ARXIV.2607.04880},
	urldate = {2026-09-16},
	publisher = {arXiv},
	author = {Ko, Dogyu and Kim, Haneul and Yeo, Chanyoung and Lee, Dowoon and Park, Taeho and Hwang, Hyoseok},
	month = jul,
	year = {2026},
}

@misc{wang_vlk:_2026,
	title = {{VLK}: {Learning} {Humanoid} {Loco}-{Manipulation} from {Synthetic} {Interactions} in {Reconstructed} {Scenes}},
	copyright = {Creative Commons Attribution 4.0 International},
	shorttitle = {{VLK}},
	note = {arXiv:2606.30645},
	doi = {10.48550/ARXIV.2606.30645},
	urldate = {2026-09-16},
	publisher = {arXiv},
	author = {Wang, Yen-Jen and Li, Jiaman and Chen, Sirui and Truong, Takara E. and Xu, Pei and Abbeel, Pieter and Duan, Rocky and Sreenath, Koushil and Kanazawa, Angjoo and Sferrazza, Carmelo and Shi, Guanya and Liu, Karen},
	month = jun,
	year = {2026},
}

@misc{ranawaka_simfoundry:_2026,
	title = {{SimFoundry}: {Modular} and {Automated} {Scene} {Generation} for {Policy} {Learning} and {Evaluation}},
	copyright = {Creative Commons Attribution Share Alike 4.0 International},
	shorttitle = {{SimFoundry}},
	note = {arXiv:2606.28276},
	doi = {10.48550/ARXIV.2606.28276},
	urldate = {2026-09-16},
	publisher = {arXiv},
	author = {Ranawaka, Nadun and Wong, Josiah and Pai, Wei-Lin and Chu, Wei-Teng and Dai, Tianyuan and Moghani, Masoud and Yin, Hang and Jiang, Yunfan and Durbano, Wesley and Huynh, Brandon and Fang, Yu and Xu, Danfei and Zhang, Ruohan and Fei-Fei, Li and Fan, Linxi and Wen, Bowen and Mandlekar, Ajay and Zhu, Yuke},
	month = aug,
	year = {2026},
}
\addtolength{\textheight}{-12cm}
\end{document}